\documentclass[sigconf]{acmart}

\usepackage{enumitem}
\usepackage{tabularx} 
\usepackage{multirow} 
\usepackage{booktabs} 
\usepackage{array}    
\usepackage{graphicx} 
\usepackage{makecell} 
\usepackage{algorithm}
\usepackage[noend]{algpseudocode}
\usepackage{fancyvrb}
\usepackage{float}
\usepackage{tcolorbox}
\usepackage{listings}
\usepackage{xcolor}

\lstdefinestyle{promptstyle}{
    basicstyle=\ttfamily\scriptsize,
    breaklines=true,
    breakatwhitespace=false, 
    frame=none,
    columns=fullflexible,
    breakindent=0pt,       
    breakautoindent=false, 
    aboveskip=0pt,  
    belowskip=0pt   
}

\AtBeginDocument{%
  }

\copyrightyear{2026}
\acmYear{2026}
\setcopyright{cc}
\setcctype{by}
\acmConference[KDD '26]{Proceedings of the 32nd ACM SIGKDD Conference on Knowledge Discovery and Data Mining V.2}{August 09--13, 2026}{Jeju Island, Republic of Korea}
\acmBooktitle{Proceedings of the 32nd ACM SIGKDD Conference on Knowledge Discovery and Data Mining V.2 (KDD '26), August 09--13, 2026, Jeju Island, Republic of Korea}
\acmDOI{10.1145/3770855.3817507}
\acmISBN{979-8-4007-2259-2/2026/08}

\begin{document}

\title{ProductWebGen: Benchmarking Multimodal Product 
 Webpage Generation}




\author{Zhihong Liu}
\authornote{Work done during an internship at Kuaishou Technology.}
\affiliation{%
  \department{School of Computer Science}\&
  \department{Zhiyuan College}
  \institution{Shanghai Jiao Tong University}
  \city{Shanghai}
  \country{China}
}
\email{lzh15841259307@sjtu.edu.cn}

\author{Siqi Kou}
\authornotemark[1]
\affiliation{%
  \institution{Shanghai Jiao Tong University}
  \city{Shanghai}
  \country{China}}
\email{happy-karry@sjtu.edu.cn}

\author{Zheng Li}
\affiliation{%
  \institution{Shanghai Jiao Tong University}
  \city{Shanghai}
  \country{China}}
\email{angra-mainyu@sjtu.edu.cn}

\author{Ye Ma}
\affiliation{%
  \institution{Kuaishou Technology}
  \city{Beijing}
  \country{China}}
\email{maye@kuaishou.com}

\author{Quan Chen}
\affiliation{%
  \institution{Kuaishou Technology}
  \city{Beijing}
  \country{China}}
\email{chenquan06@kuaishou.com}

\author{Peng Jiang}
\affiliation{%
  \institution{Kuaishou Technology}
  \city{Beijing}
  \country{China}}
\email{jiangpeng@kuaishou.com}

\author{Kai Yu}
\affiliation{%
  \institution{Shanghai Jiao Tong University}
  \city{Shanghai}
  \country{China}}
\email{kai.yu@sjtu.edu.cn}

\author{Zhijie Deng}
\authornote{Corresponding author.}
\affiliation{%
  \institution{Shanghai Jiao Tong University}
  \city{Shanghai}
  \country{China}}
\email{zhijied@sjtu.edu.cn}

\renewcommand{\shortauthors}{Zhihong Liu et al.}


\begin{abstract}
Crafting a product display webpage from a source product image, along with layout and visual content instructions, holds significant practical value for domains such as marketing, advertising, and E-commerce. 
Intuitively, this task demands strict visual consistency across product displays and high-fidelity instruction following to jointly generate renderable HTML code. 
These requirements on controllability and instruction-following are closely aligned with the core features of advanced multimodal generative models, such as image editing models and unified models (UMs). 
To this end, this paper introduces ProductWebGen to systematically benchmark the product webpage generation capacities of these models. 
We organize ProductWebGen with 500 test samples covering 13 product categories; each sample consists of a source image, a visual content instruction, and a webpage instruction. 
The task is to generate a product showcase webpage including multiple consistent images in accordance with the source image and instructions. 
Given the mixed-modality input-output nature of the task, we design and systematically compare two workflows for evaluation---one uses large language models (LLMs) and image editing models to separately generate HTML code and images (editing-based),
while the other relies on a single UM to generate both, with image generation conditioned on the preceding multimodal context (UM-based).
Empirical results show that editing-based approaches achieve leading results in webpage instruction following and content appeal, while UM-based ones may display more advantages in fulfilling visual content instructions. 
We also construct a supervised fine-tuning (SFT) dataset, ProductWebGen-1k, with 1,000 groups of real product images and LLM-generated HTML code. 
We verify its effectiveness on the open-source UM BAGEL. 
The benchmark, training dataset, and inference code are publicly  available at \url{https://github.com/SJTU-DENG-Lab/ProductWebGen}.
\end{abstract}




\maketitle

\section{Introduction}

\begin{figure*}[t]
    \centering
    \includegraphics[width=\textwidth]{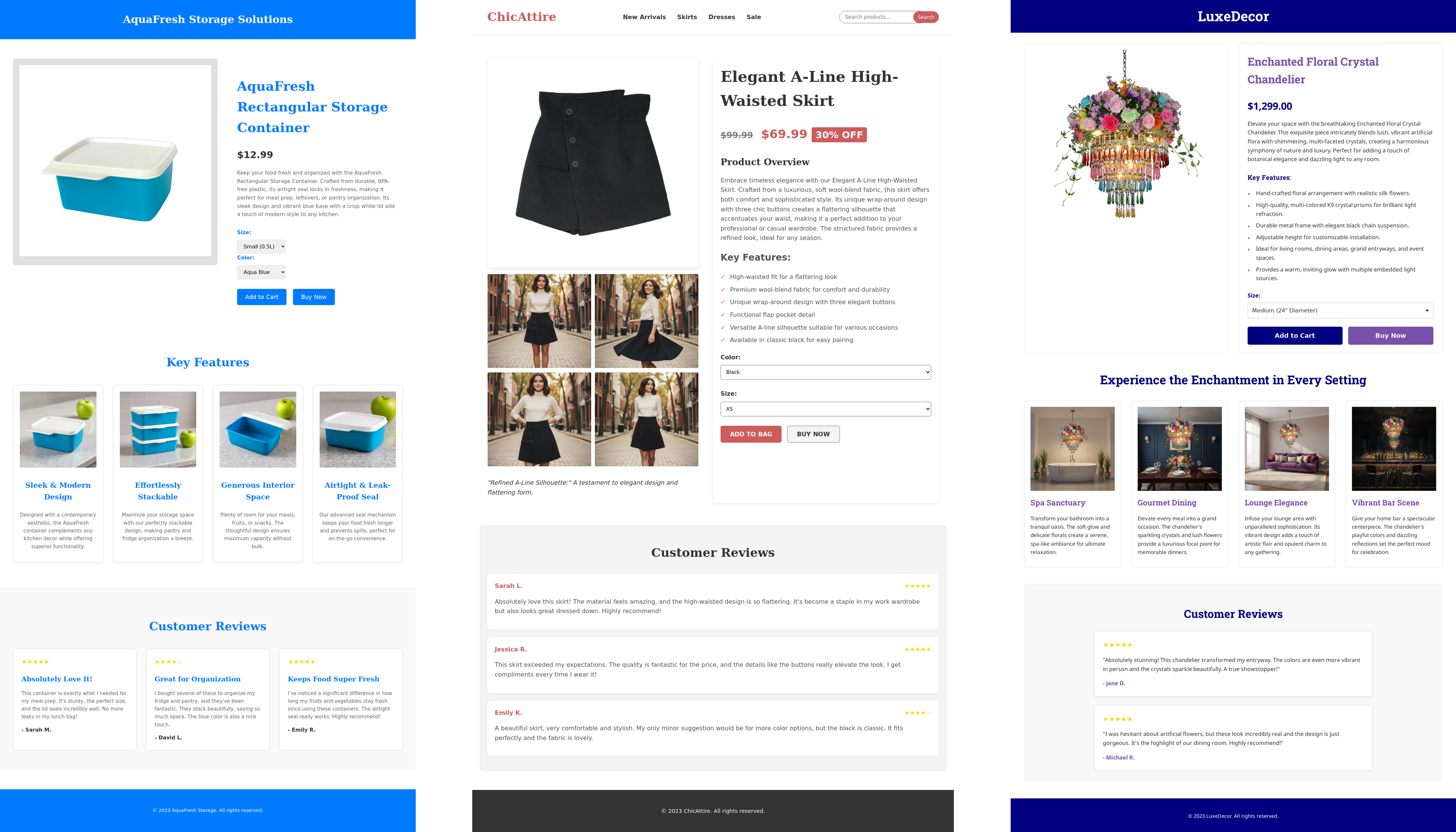}
    \caption{Illustrative examples of webpages generated on the ProductWebGen benchmark. These examples showcase the complex, multimodal nature of the task. It requires models to jointly generate renderable HTML code for the webpage layout and content, and multiple, visually consistent images for the product showcase.}
    \label{fig:teaser}
\end{figure*}

Generating a product display webpage from a source product image and accompanying layout/visual content instructions offers substantial practical value for fields including marketing, advertising, and e-commerce.
Unlike the simple image generation or editing tasks, this task presents substantial challenges for multimodal generative models. Specifically, it requires (1) strict visual consistency, ensuring that multiple product display images maintain coherence, and (2) high-fidelity instruction following, necessary for generating renderable HTML code and text that precisely adheres to layout and style specifications.

These specific requirements align closely with the core capabilities of state-of-the-art multimodal generative models. Advanced image editing models, such as FLUX.1 Kontext~\cite{batifol2025flux} and Qwen-Image-Edit~\cite{wu2025qwen}, are specialized in controlled and consistent editing, which is essential for maintaining visual consistency among multiple product display images. Concurrently, there has been growing interest in conjoining image understanding and generation within unified models (UMs) for mixed-modality generation~\cite{pan2025transfer,zhou2025transfusion,chen2025blip3,wang2024emu3,xie2024show,wu2025omnigen2,xie2025show}, with BAGEL~\cite{deng2025emerging} and Gemini-2.5-Flash-Image~\cite{google_gemini2_5_flash_image_2025} as popular examples.

This paper introduces the ProductWebGen benchmark to systematically evaluate the ability of existing multimodal generative models to fulfill the practical requirements of product webpage generation.
Specifically, ProductWebGen includes 500 carefully curated samples spanning 13 distinct product categories, where each sample consists of a carefully designed user instruction and a source product image. 
As shown in Figure \ref{fig:benchmark sample example}, the user instruction contains two parts for controlling generation: a \emph{visual content instruction} which imposes consistency requirements among the generated images, and \emph{webpage instructions} which specify the layout, style, and textual content of the webpage. 
Compared to prior multimodal understanding or generation benchmarks~\cite{yue2024mmmu,ghosh2023geneval,niu2025wise}, ProductWebGen not only requires basic knowledge (e.g., the use of existing CSS styles) and generation capabilities of HTML, but also entails the ability to generate images given long, multimodal contexts. 

Compared to HTML code, the generation of images on the webpage poses higher challenges in practice. 
According to how the images are generated, we design two baselines (see Figure~\ref{fig:workflow}). 

One is the \emph{editing-based} approach---a large language model (LLM) is first invoked to produce a set of textual descriptions for the images to generate, which, in conjunction with the source image, are then fed into image editing models to produce the images. 
The other is the \emph{UM-based (HTML)} approach---we let the UM generate the images given an image-HTML interleaved context, which is expected to enjoy better image consistency.
Considering that the HTML code can be long and raise long-context challenges, we also try to replace the HTML code with textual descriptions generated by the UM itself during the interleaved generation of the images, giving rise to the \emph{UM-based} approach. 
In our empirical studies, we combine leading LLMs, including Gemini-2.5-Flash~\cite{comanici2025gemini}, GPT-4o~\cite{hurst2024gpt}, Grok-4~\cite{xai_grok4_2025}, and Claude-Sonnet-4~\cite{anthropic_claude_sonnet4_2025}, with specialized image editing models like Qwen-Image-Edit~\cite{wu2025qwen} and FLUX.1-Kontext~\cite{batifol2025flux} to specify \emph{editing-based} approaches. 
For \emph{UM-based (HTML)} and \emph{UM-based} ones, we evaluate three open-source models BAGEL~\cite{deng2025emerging}, Ovis-U1~\cite{wang2025ovis}, and OmniGen2~\cite{wu2025omnigen2}, as well as a closed-source model Gemini-2.5-Flash-Image~\cite{google_gemini2_5_flash_image_2025}. 
We leverage LLM-as-a-judge~\cite{zheng2023judging} to rate the generated webpage from multiple aspects like instruction following and visual appeal. 
Our key findings are: 

\begin{itemize}[left=0pt]
    \item The \emph{UM-based} approach with Gemini-2.5-Flash-Image shows the best overall performance.
    \item \emph{Editing-based} approaches excel at webpage instruction following and image perception quality, while \emph{UM-based} ones can be superior in visual content consistency.
    \item \emph{UM-based} approaches are usually better than \emph{UM-based (HTML)} ones in visual content consistency, which implies that complex HTML code within the context can impair the visual content instruction following ability of UMs.
    \item There is a significant performance gap between open-source UMs and the closed-source Gemini-2.5-Flash-Image.
\end{itemize}

Furthermore, we construct a supervised fine-tuning (SFT) dataset, ProductWebGen-1k, and verify its effectiveness on BAGEL.
We observe significant performance improvement: +22.3\% in visual content instruction following and +65.0\% in webpage instruction following.

\section{The ProductWebGen Benchmark}

ProductWebGen requires the model to generate webpages with rich visual content for product showcase, according to a source product image and a user instruction. 
Overall, ProductWebGen contains 500 curated test samples spanning 13 product categories, including \textit{food, apparel, beauty, household supplies, digital products, appliances, baby products, office supplies, pet supplies, furniture, sports, jewelry, and kitchenware}. 
We describe more details below.

\subsection{Data Curation}
\label{sec:test set curation}

\begin{figure}[t]
    \centering
    \includegraphics[width=1.0\linewidth]{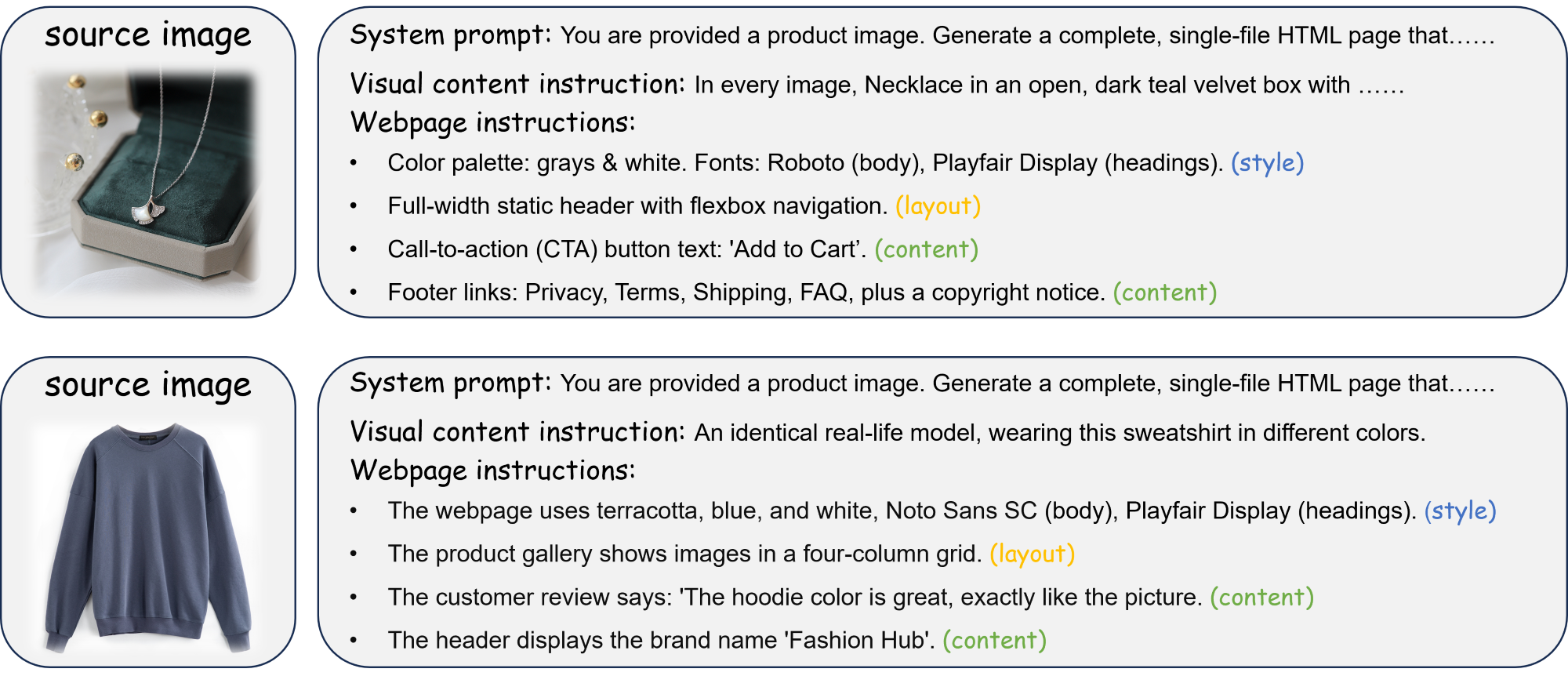}
    \caption{Two test samples from ProductWebGen, which both consist of a source image, a webpage instruction, and a visual content instruction.
    The system prompt is shared across samples.
    }
    \label{fig:benchmark sample example}

\end{figure}

As illustrated in Figure~\ref{fig:benchmark sample example}, the test sample in ProductWebGen comprises two parts: a source product image and a user instruction. The instruction comprises three components: \textit{system prompt}, \textit{visual content instruction}, and \textit{webpage instruction}. \textit{System prompt} is identical across all samples, which specifies the task, I/O formats, etc. The complete system prompt can be found in Appendix \ref{app:visual content instruction generation prompt}. \textit{Visual content instruction} asks the model to maintain consistency among the generated images. The consistency boils down to four categories: background consistency, character consistency, watermark consistency, and perspective coherence. 
\textit{Webpage instruction} specifies the requirements for the style, layout, and content of the webpage. 

We crawl source product images from the Internet in compliance with legal regulations. 
For visual content instructions, we randomly select one from the four aforementioned consistency categories and prompt LLMs to generate detailed instructions according to the source product image. 

For webpage instructions, to ensure their validity, we first use LLMs to generate diverse seed HTML webpages for the product images, from which the instructions regarding style, layout, and content are extracted.
See Appendix \ref{app:visual content instruction generation prompt} for details of the used prompts.

\subsection{Metrics}

\label{section:metric}

Given the lack of quantitative metrics for evaluating multimodal webpage quality, we define metrics based on LLM-as-a-judge ~\cite{zheng2023judging} following common practice (see Appendix \ref{app:metric prompt} for the prompts). 
We defer the study on the alignment of these metrics with human evaluations to Section \ref{sec:human evaluation}.

\emph{\textbf{Webpage Instruction Following (WIF)}} evaluates whether the generated HTML code follows the clauses in the webpage instruction regarding style, layout, and content. 
    The LLM accepts both the HTML code and the webpage instruction as input and outputs 1 (following) or 0 (not following) for each clause. We report the average score over these clauses.  

\emph{\textbf{Webpage Design Quality (WDQ)}} evaluates the style and layout of the webpage, including its visual hierarchy, layout, color, and overall aesthetic appeal. We input the screenshot of the rendered webpage into a multimodal LLM (MLLM) and get a score between 0 and 10.

\emph{\textbf{Webpage Content Appeal (WCA)}} evaluates the effectiveness and appeal of the webpage content, considering promotional language, details on after-sales service, authentic customer reviews, etc. We input the webpage screenshot into an MLLM and get a score between 0 and 10.

\emph{\textbf{Visual Content Instruction Following (VCIF)}} evaluates how well the generated images follow the visual content instruction. An MLLM accepts the source image, all generated images, and the visual content instruction as input and outputs a raw score between 0 and 5, which is then linearly mapped to a 0 -- 10 scale.

\emph{\textbf{Image Perception Quality (IPQ)}} evaluates the visual authenticity and naturalness of the generated image. Following VIEScore ~\cite{ku2023viescore}, we input a generated image to an MLLM and get a score between 0 and 10. The average over all the generated images for a webpage is reported.

The first three metrics are webpage-related, while the following two are image-related.

\subsection{Design of Evaluation Approaches}

\label{sec:benchmarking method}

\begin{figure*}[t]
    \centering
    \includegraphics[width=1.0\linewidth]{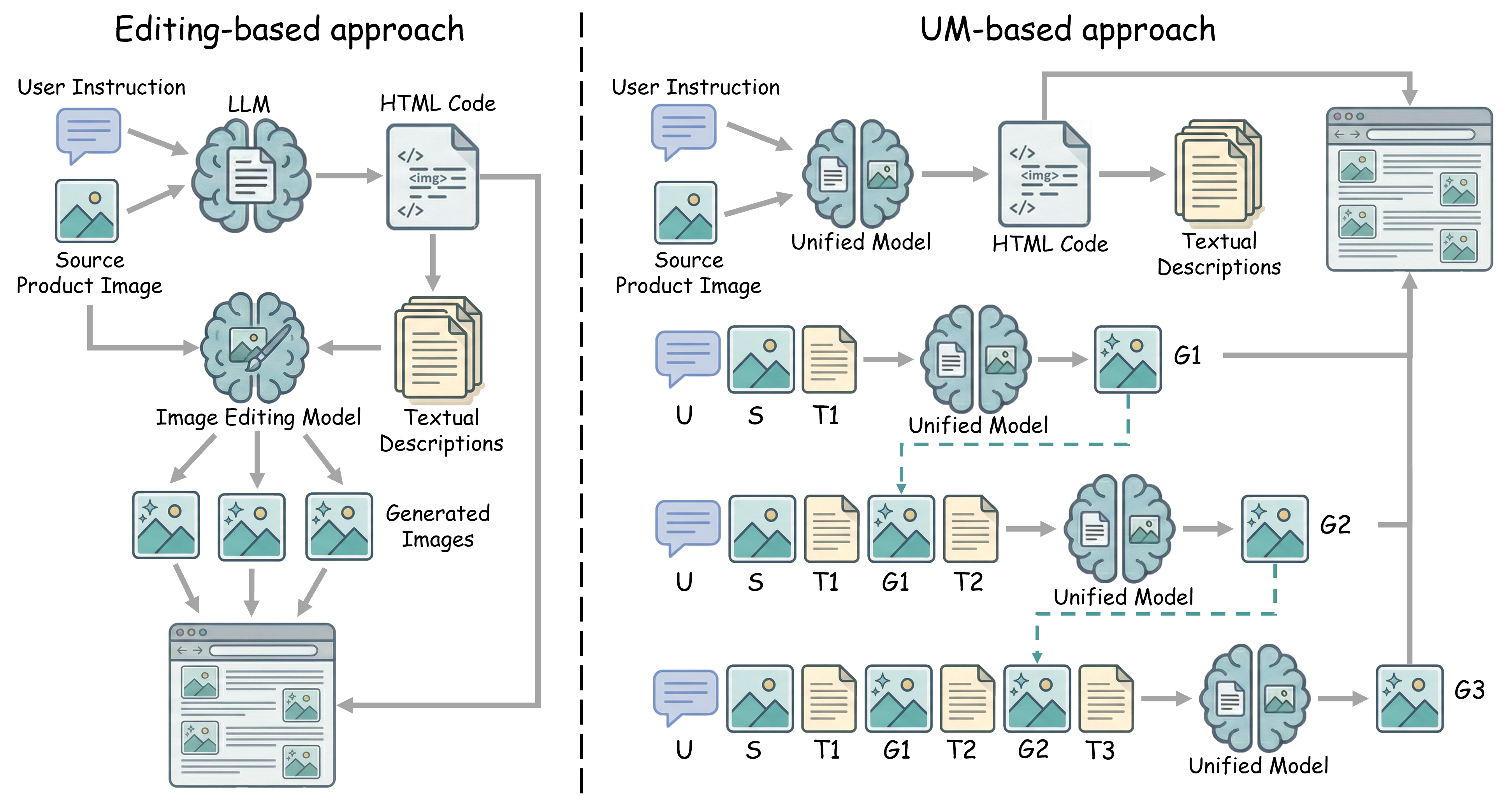}
    \caption{Two baseline approaches for ProductWebGen. 
    \emph{Editing-based} approaches produce images with an image editing model, based on the source product image and LLM-generated textual descriptions for the images to display. \emph{UM-based} approaches use multimodal context to inform image generation. We denote User Instruction, Source Product Image, Textual Description, and Generated Image as U, S, T, and G, respectively.}
    \label{fig:workflow}
\end{figure*}

According to the capacities of existing multimodal generative models, we design two kinds of baselines. The comparison between them is displayed in Figure~\ref{fig:workflow}.

\textbf{\emph{Editing-based} approach} disentangles the generation of HTML code and images for simplicity. 
It leverages the fact that the images to be displayed are usually edition variants of the source image. Specifically, the approach generates both the HTML code and the \emph{descriptions for the images to be generated} in a single LLM call by embedding the descriptions directly within the \texttt{alt} attribute of the \texttt{<img>} tags in the HTML. These descriptions, paired with the source image, are then fed into an image editing model to produce the final display images.

\textbf{\emph{UM-based} approach} can, ideally, produce interleaved HTML and images in a sequential manner, with flexibly determined modality transition. 
However, our preliminary tests revealed that existing UMs face challenges in autonomously switching between code and image generation. Furthermore, even when we manually enforce this transition—by inserting images into the context and prompting the model to resume code generation—we observed issues such as truncated code, incorrect image counts, or incomplete webpage structures.
To mitigate this, we first have UMs generate the entire HTML code, with descriptions for image generation embedded in the \texttt{alt} attributes of \texttt{<img>} tags, similar to the \emph{editing-based} approach. For image generation, we first attempt to generate images from an interleaved image-HTML context (i.e., the image is generated conditioning on all the preceding elements of the corresponding \texttt{<img>} tag), yielding the \emph{UM-based (HTML)} baseline.
Given the potential long-context challenge posed by the combined HTML code and images, we also explore an alternative context definition that interleaves images with the aforementioned descriptions, resulting in the default \emph{UM-based} approach.

\section{Results and Analysis}

\begin{table*}[t]
\centering
\caption{Results of \emph{editing-based} approaches. VCIF and IPQ evaluate visual instruction following and image quality. WIF, WDQ, and WCA evaluate webpage instruction following, design quality, and content appeal. The best result for every metric is highlighted in bold.}
\label{tab:editing-based}
\small
\begin{tabularx}{\textwidth}{@{} l l *{5}{>{\centering\arraybackslash}X} @{}}
\toprule
\multirow{3}{*}{\textbf{LLM}} & \multirow{3}{*}{\textbf{Image Editing Model}} & \multicolumn{2}{c}{\textbf{Image-related}} & \multicolumn{3}{c}{\textbf{Webpage-related}} \\
\cmidrule(lr){3-4} \cmidrule(lr){5-7}
& & \textbf{VCIF (0-10)} & \textbf{IPQ (0-10)} & \textbf{WIF (0-1)} & \textbf{WDQ (0-10)} & \textbf{WCA (0-10)} \\
\midrule
\multirow{2}{*}{Gemini-2.5-Flash} & Qwen-Image-Edit & \textbf{6.45} &\textbf{8.00} & 0.81 & 7.95 & 7.44 \\
& FLUX.1-Kontext & 5.75 &7.66 & 0.81 & 7.92 & 7.42 \\
\midrule
\multirow{2}{*}{Gpt-4o} & Qwen-Image-Edit & 5.97 &7.86 & 0.78 & 7.72 & 6.37 \\
& FLUX.1-Kontext & 4.70 &7.48 & 0.78 & 7.64 & 6.28 \\
\midrule
\multirow{2}{*}{Claude-sonnet-4} & Qwen-Image-Edit & 5.74 &\textbf{8.00} & 0.84 & 7.93 & \textbf{7.45} \\
& FLUX.1-Kontext & 4.94 &7.65 & 0.84 & \textbf{7.98} & 7.42 \\
\midrule
\multirow{2}{*}{Grok-4} & Qwen-Image-Edit & 5.30 &\textbf{8.00} & \textbf{0.87} & 7.93 & 7.21 \\
& FLUX.1-Kontext & 4.73 &7.68 & \textbf{0.87} & 7.86 & 7.15 \\
\bottomrule
\end{tabularx}
\end{table*}

\begin{table*}[t]
\centering
\caption{Results of \emph{UM-based} approaches. VCIF and IPQ evaluate visual instruction following and image quality. WIF, WDQ, and WCA evaluate webpage instruction following, design quality, and content appeal. 
}
\label{tab:UM-based}
\small
\begin{tabularx}{\textwidth}{@{} l l *{5}{>{\centering\arraybackslash}X} @{}}
\toprule
& \multirow{3}{*}{\textbf{Unified Model}}& \multicolumn{2}{c}{\textbf{Image-related}} & \multicolumn{3}{c}{\textbf{Webpage-related}} \\
\cmidrule(lr){3-4} \cmidrule(lr){5-7}
& & \textbf{VCIF (0-10)} & \textbf{IPQ (0-10)} & \textbf{WIF (0-1)} & \textbf{WDQ (0-10)} & \textbf{WCA (0-10)} \\
\midrule
\multirow{4}{*}{\emph{UM-based (HTML)}} 
  & BAGEL                     & 4.22 & 5.27 & 0.40 &6.96  & 5.45 \\
  & Ovis-U1                   & 5.06 & 2.83 & 0.37 &6.39  & 4.73 \\
  & OmniGen2                  & 3.91 & 2.05 & 0.42 &6.59  & 4.54 \\
  & Gemini-2.5-Flash-Image    & 7.24 & 8.16 & \textbf{0.84} & \textbf{7.94} & 7.27 \\
\midrule
\multirow{4}{*}{\emph{UM-based}} 
  & BAGEL                     & 5.84 & 5.43 & 0.40 & 7.26 & 5.61 \\
  & Ovis-U1                   & 6.27 & 4.17 & 0.37 & 6.31 & 5.02 \\
  & OmniGen2                  & 6.56 & 5.49 & 0.42 & 6.56 & 5.04 \\
  & Gemini-2.5-Flash-Image    & \textbf{8.15} & \textbf{8.35} & \textbf{0.84} & 7.92 & \textbf{7.31} \\
\bottomrule
\end{tabularx}
\end{table*}

\subsection{Model Setup}

\textbf{\emph{Editing-based} approach.} We select four prevalent LLMs, i.e., Gemini-2.5-Flash ~\cite{comanici2025gemini}, GPT-4o ~\cite{hurst2024gpt}, Grok-4 ~\cite{xai_grok4_2025}, and Claude-Sonnet-4 ~\cite{anthropic_claude_sonnet4_2025}, and two advanced image editing models, Qwen-Image-Edit ~\cite{wu2025qwen} and FLUX.1-Kontext ~\cite{batifol2025flux}, evaluating their combinations. 

\textbf{\emph{UM-based} approach.} We evaluate three open-source UMs, i.e., BAGEL ~\cite{deng2025emerging}, Ovis-U1 ~\cite{wang2025ovis}, and OmniGen2 ~\cite{wu2025omnigen2}, and one state-of-the-art closed-source model Gemini-2.5-Flash-Image ~\cite{google_gemini2_5_flash_image_2025} (a.k.a., nano-banana). BAGEL adopts two transformer experts for multimodal understanding and generation while sharing self-attention for information fusion. 
Ovis-U1 and OmniGen2 use multimodal LLMs (MLLMs) to embed multimodal contexts, and use the embeddings as conditions for a diffusion decoder to generate images. 


\textbf{LLM-as-a-judge.} Recent study ~\cite{wataoka2024self} reveals the potential for self-preference bias when using LLMs-as-a-judge, particularly when the judge model is related to the systems being evaluated. To mitigate such conflicts and ensure a rigorous, objective evaluation, we select two powerful, independent third-party models to serve as our judges. Specifically, for the webpage-related metrics, we employ GLM-4.5 ~\cite{zeng2025glm} and GLM-4.5V ~\cite{v2025glm}. For VCIF metric, we employ Qwen3-VL-235B-A22B-Instruct ~\cite{qwen3vl}. For IPQ metric, we follow VIEScore ~\cite{ku2023viescore}, which utilizes GPT-4o. The use of GPT-4o does not introduce self-preference bias, as it is not used to generate images.


Details on our experimental setup are provided in Appendix \ref{app:experiment detail}.

\subsection{Quantitative Results}

\label{sec:Automatic Evaluation}

\begin{table*}[t]
\centering
\caption{Results of \emph{UM-based} approaches based on the HTML code and textual descriptions generated by Gemini-2.5-Flash. VCIF and IPQ evaluate visual instruction following and image quality. WIF, WDQ, and WCA evaluate webpage instruction following, design quality, and content appeal. 
}
\label{tab:HTML_controlled}
\small
\begin{tabularx}{\textwidth}{@{} l l *{5}{>{\centering\arraybackslash}X} @{}}
\toprule
 & \multirow{3}{*}{\textbf{Unified Model}} & \multicolumn{2}{c}{\textbf{Image-related}} & \multicolumn{3}{c}{\textbf{Webpage-related}} \\
\cmidrule(lr){3-4} \cmidrule(lr){5-7}
& & \textbf{\begin{tabular}[c]{@{}c@{}}VCIF \\ (0-10)\end{tabular}} & \textbf{\begin{tabular}[c]{@{}c@{}}IPQ \\ (0-10)\end{tabular}} & \textbf{\begin{tabular}[c]{@{}c@{}}WIF \\ (0-1)\end{tabular}} & \textbf{\begin{tabular}[c]{@{}c@{}}WDQ \\ (0-10)\end{tabular}} & \textbf{\begin{tabular}[c]{@{}c@{}}WCA \\ (0-10)\end{tabular}} \\
\midrule
\multirow{3}{*}{\emph{UM-based (HTML)}} 
  & BAGEL & 2.32 &3.93 & 0.81 & 7.91 & 7.33 \\
  & Ovis-U1    & 4.73 &3.13 & 0.81 & 7.84 & 7.28 \\
  & OmniGen2  & 0.72 &0.95 & 0.81 & 7.95 & 7.41 \\
\midrule
\multirow{3}{*}{\emph{UM-based}} 
  & BAGEL    & 6.64 &6.07 & 0.81 & 7.86 & 7.38 \\
  & Ovis-U1    & \textbf{6.92} &\textbf{6.57} & 0.81 & 7.95 & 7.45 \\
  & OmniGen2  & \textbf{6.92} &5.98 & 0.81 & 7.94 & 7.47 \\
\bottomrule
\end{tabularx}
\end{table*}

\begin{table}[t]
\centering
\caption{IPQ breakdown by image generation order for the \emph{UM-based} results in Table~\ref{tab:HTML_controlled}. Img $i$ IPQ denotes the quality of the $i$-th generated image.}
\label{tab:IPQ breakdown}
\small
\setlength{\tabcolsep}{2pt}
\resizebox{\columnwidth}{!}{%
\begin{tabular}{@{} l cccc @{}}
\toprule
\textbf{Method} & \textbf{Img1 IPQ} & \textbf{Img2 IPQ} & \textbf{Img3 IPQ} & \textbf{Img4 IPQ} \\
\midrule
BAGEL & 6.63 & 6.12 & 5.95 & 5.58 \\
Ovis-U1 & 7.36 & 7.10 & 6.27 & 5.55 \\
OmniGen2 & 6.97 & 6.75 & 5.93 & 4.27 \\
Gemini-2.5-Flash-Image & 8.29 & 8.47 & 8.23 & 8.42 \\
\bottomrule
\end{tabular}%
}
\end{table}

\begin{table}[t]
    \centering
    \setlength{\tabcolsep}{2.5pt}
    
    \caption{Performance breakdown of Visual Content Instruction Following (VCIF) across different instruction types. The \emph{Editing-based} column reports the average score of different models. The best result for each type is highlighted in bold.}
    \label{tab:vcif_breakdown}
    
    \resizebox{\linewidth}{!}{
        \begin{tabular}{lccccc}
            \toprule
            \multirow{2}[3]{*}{\textbf{Instruction Type}} & \textbf{\emph{Editing-based}} & \multicolumn{4}{c}{\textbf{\emph{UM-based}}} \\
            \cmidrule(lr){3-6} 
            
             &  & \textbf{Gemini-2.5-} & \textbf{BAGEL} & \textbf{Ovis-U1} & \textbf{OmniGen2} \\
             & & \textbf{Flash-Image} & & & \\
             
            \midrule
            Character Consistency & 4.96 & 8.40 & 7.79 & 8.52 & \textbf{8.93} \\
            Watermark Consistency & 4.76 & \textbf{8.10} & 2.45 & 3.81 & 5.58 \\
            Background Consistency & 7.56 & \textbf{9.32} & 5.03 & 6.87 & 7.28 \\
            Perspective Coherence & 4.09 & \textbf{5.05} & 3.70 & 2.51 & 1.56 \\
            \bottomrule
        \end{tabular}
    }
\end{table}

We present the quantitative results of the \emph{editing-based} and \emph{UM-based} approaches in Table \ref{tab:editing-based} and Table \ref{tab:UM-based}, respectively. 
We summarize our key findings as follows:

\textbf{The \emph{UM-based} approach with Gemini-2.5-Flash-Image shows the best overall performance.} As shown, Gemini-2.5-Flash-Image achieves the highest scores on the visual content instruction following, image perception quality, with other metrics also near the best. This likely stems from its strong code generation capabilities inherited from Gemini-2.5-Flash, as well as its powerful ability for interleaved text and image generation. 

\textbf{\emph{Editing-based} approach performs better on webpage-related metrics.} The combination of Claude-Sonnet-4 and the two image editing models achieves the highest scores on webpage design quality and webpage content appeal. Grok-4 obtains top scores on the webpage instruction following. In contrast, \emph{UM-based} approaches, except for Gemini-2.5-Flash-Image, perform mostly poorly on webpage-related metrics. This can be attributed to the \emph{editing-based} approach leveraging leading LLMs to generate HTML code.

\textbf{\emph{UM-based} approach can be superior in visual content consistency, but open-source UMs lag behind.} The closed-source UM Gemini-2.5-Flash-Image achieves a visual content instruction following score of 8.15, exceeding the best of the \emph{editing-based} approach (6.45) by 26.4\%. 
This advantage stems from the use of previously generated images and descriptions to guide new image generation, which helps maintain consistency across multiple images. In contrast, the \emph{editing-based} approach relies solely on the source image and descriptions when generating.
However, the open-source UMs significantly lag in both visual content instruction following and image quality. 
To investigate the cause, we conduct a study using the Gemini-2.5-Flash to generate HTML code (as well as textual descriptions in \texttt{alt}) and use UMs for interleaved image generation. 
As shown in Table \ref{tab:HTML_controlled}, the visual content instruction following score of BAGEL increases from 5.84 to 6.64. 
This suggests that one of the causes for the original gap is that the open-source UMs fail to generate sufficiently good descriptions for the images to generate. This conclusion is further supported by the length of the generated alt text: Gemini-2.5-Flash produces 85 alt-text tokens on average, whereas open-source UMs produce at most 20.
Nevertheless, a considerable gap still remains compared to Gemini-2.5-Flash-Image.
Table \ref{tab:IPQ breakdown} breaks down the IPQ results of the \emph{UM-based} setting in Table \ref{tab:HTML_controlled} by image generation order. Img1 IPQ reflects pure generation ability, since it is not yet affected by previously generated images.
Open-source UMs yield lower Img1 IPQ than Gemini-2.5-Flash-Image, indicating a fundamental gap in image generation capability. Moreover, open-source UMs show clear IPQ degradation from Img1 to Img4, whereas Gemini-2.5-Flash-Image remains stable. This degradation lowers image quality and weakens cross-image consistency, contributing to the remaining VCIF gap.

\textbf{HTML code within the context impairs the visual content instruction following ability of UMs.} As shown in Table \ref{tab:UM-based}, the \emph{UM-based (HTML)} approach yields lower performance in visual content instruction following across all unified models compared to the \emph{UM-based} approach. Unlike natural language, HTML contains extensive elements that lack semantic information. The semantic content resides in the image descriptions, which, yet, occupy only a small fraction of the context. Consequently, the \emph{UM-based (HTML)} approach often overlooks critical information and suffers from degraded performance in visual content instruction following. 

\textbf{Performance breakdown among different types of visual content instruction.} As shown in Table \ref{tab:vcif_breakdown}, the \emph{UM-based} approach achieves the best results for all types of visual content instruction. Specifically, UMs demonstrate a significant advantage in character consistency and watermark consistency. For background consistency, the performance gap between the two approaches is relatively small. Regarding perspective coherence, models from both approaches yield universally low scores, with the highest score reaching only 5.05. This indicates that this instruction type presents a universal challenge to current models.

\textbf{The key findings generalize consistently across product categories.} To examine whether the key findings hold across the 13 product categories, we conduct category-level comparisons for three key conclusions: (1) Gemini-2.5-Flash-Image outperforms the \emph{editing-based} approach in VCIF; (2) Gemini-2.5-Flash-Image outperforms open-source UMs in VCIF and IPQ; and (3) the \emph{editing-based} approach outperforms open-source UMs in WIF, WDQ, and WCA. We select representative systems for each approach: Gemini-2.5-Flash-Image, BAGEL, and Gemini-2.5-Flash + Qwen-Image-Edit, respectively. As shown in Appendix Table~\ref{tab:category_generalization}, all corresponding performance margins are strictly positive in every category, confirming that our key conclusions generalize uniformly across product types.

\subsection{Qualitative Results}

In this section, we provide a qualitative demonstration for some of the key findings in Section \ref{sec:Automatic Evaluation} and conduct a detailed analysis with specific examples.

\begin{figure*}[t]
    \centering
    \includegraphics[width=1.0\linewidth]{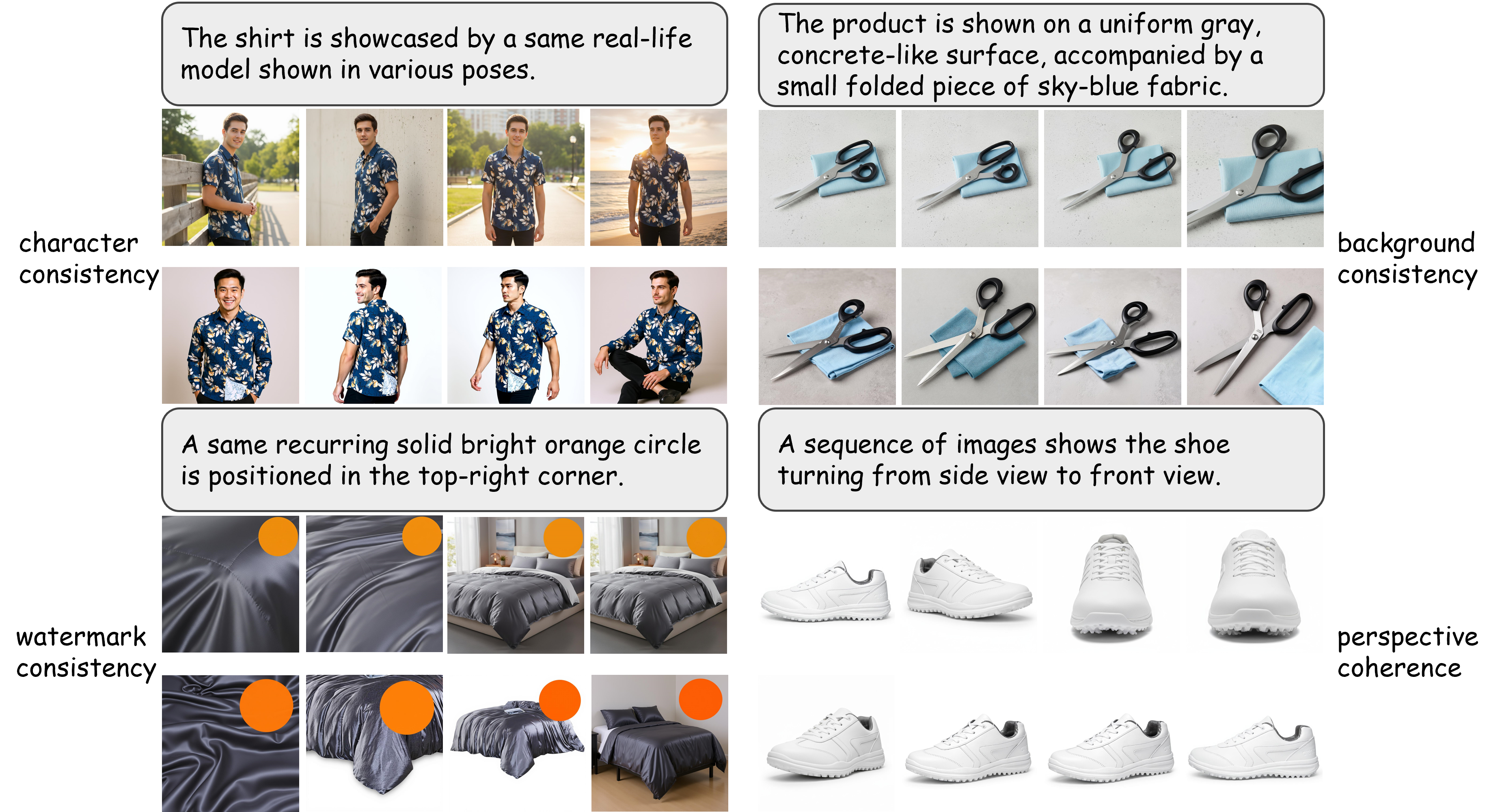}
    \caption{A comparison of the \emph{editing-based} approach (Gemini-2.5-Flash + Qwen-Image-Edit, bottom row) and the \emph{UM-based} approach (Gemini-2.5-Flash-Image, top row) for visual content instruction following. 
    The types of visual content instructions from left to right are, respectively: character consistency, background consistency, watermark consistency, and perspective coherence. 
    The \emph{UM-based} approach achieves better performance across all types of visual content instructions.}
    \label{fig:visual content instruction Qualitative results}
\end{figure*}

\textbf{Advantage of \emph{UM-based} approach in visual content instruction following.} Figure \ref{fig:visual content instruction Qualitative results} compares images from the \emph{UM-based} approach based on Gemini-2.5-Flash-Image (top row) and the \emph{editing-based} approach with Gemini-2.5-Flash + Qwen-Image-Edit (bottom row) for 4 types of visual content instruction. 
It is visually apparent that the \emph{UM-based} approach adheres more closely to the visual content instruction, as reflected in details such as the same human model across images, the identical fabric and surface textures under the scissors, the uniform watermark color and size, and the shoe's varied display angles. 

\begin{figure}[t]
    \centering
    \includegraphics[width=1.0\linewidth]{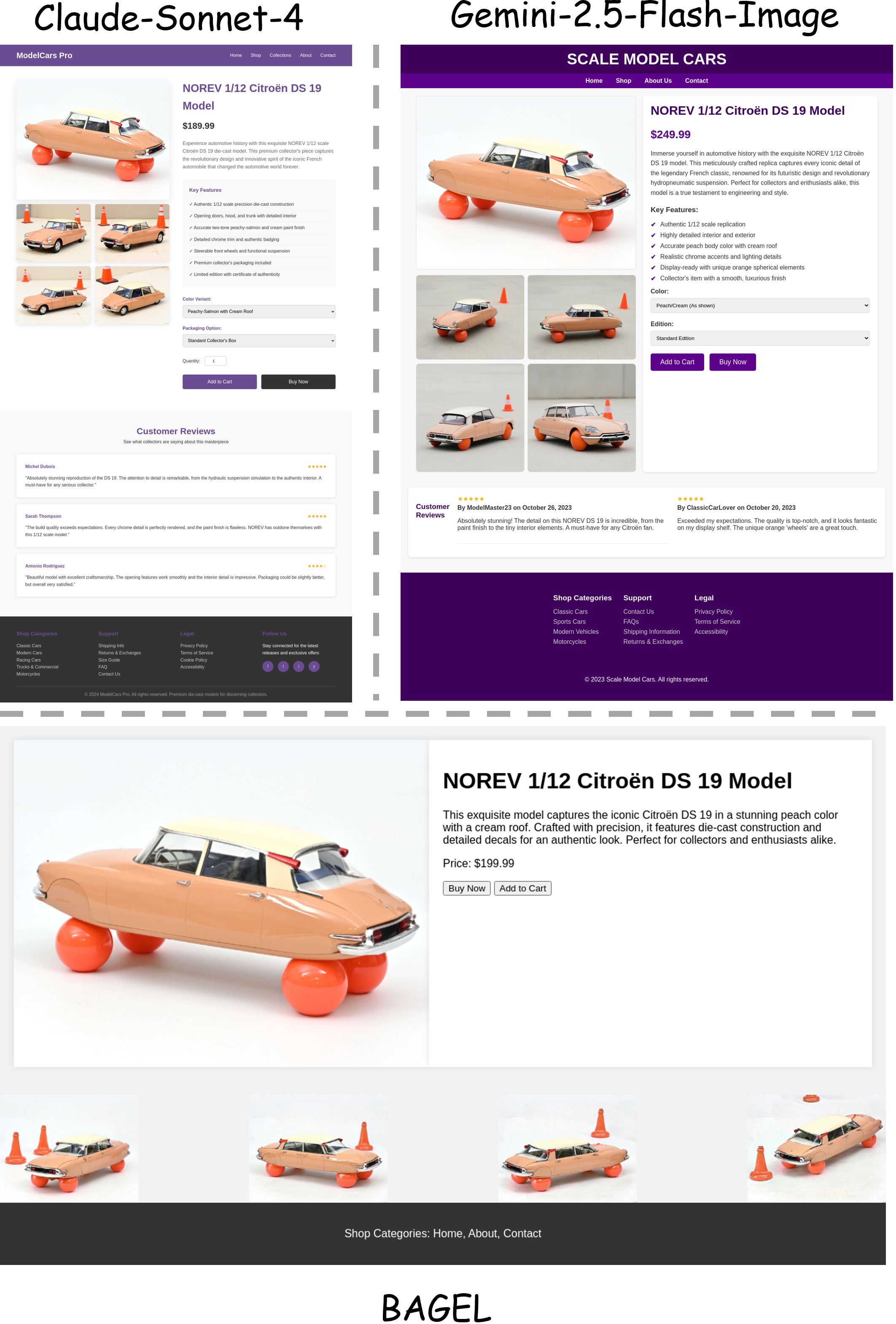}
    \caption{A comparison of the \emph{editing-based} approach (Claude-Sonnet-4 + Qwen-Image-Edit) and \emph{UM-based} approaches (Gemini-2.5-Flash-Image and BAGEL) for webpage design quality and webpage content appeal.}
    \label{fig:webpage Qualitative results}
\end{figure}

\begin{figure}[t]
    \centering
    \includegraphics[width=1.0\linewidth]{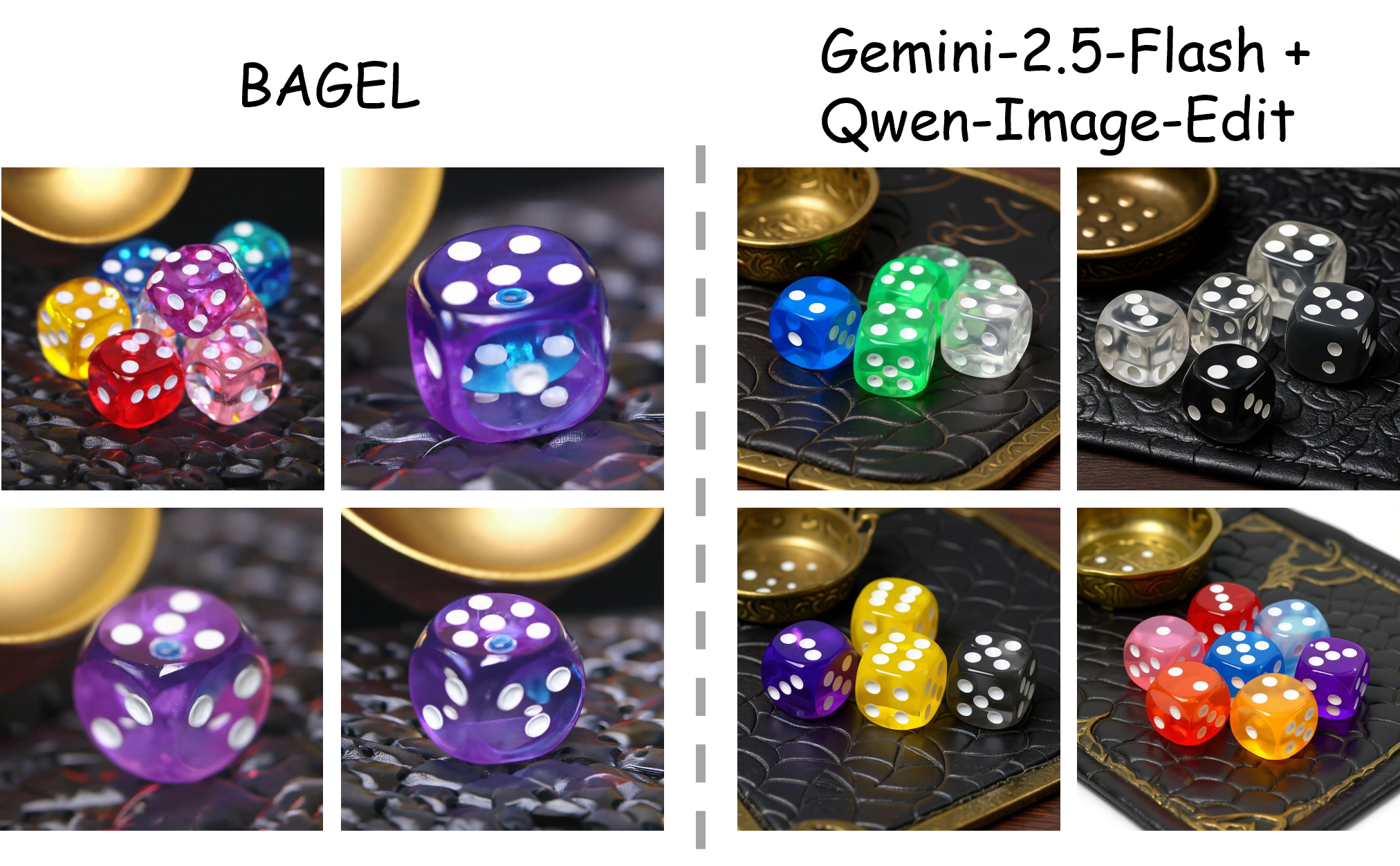}
    \caption{A comparison of the \emph{editing-based} approach and the \emph{UM-based} approach for image perception quality. The dice generated by BAGEL exhibit obvious deformation, and the dot arrangement in the images is also unreasonable. In contrast, the images generated by the \emph{editing-based} approach (Gemini-2.5-Flash + Qwen-Image-Edit) are of higher quality.}
    \label{fig:image quality Qualitative results}
\end{figure}

\textbf{Advantage of \emph{editing-based} approach on webpage-related metrics.} Figure \ref{fig:webpage Qualitative results} compares webpages generated by the \emph{editing-based} approach (Claude-Sonnet-4 + Qwen-Image-Edit) and two UMs (Gemini-2.5-Flash-Image and BAGEL) for the same test case. 
As shown, the webpages from the \emph{editing-based} approach and Gemini-2.5-Flash-Image are of similar quality, both clearly superior to BAGEL. 
Furthermore, the \emph{editing-based} approach can produce images with higher visual quality and more details due to the use of SOTA editing models like Qwen-Image-Edit and FLUX.1-Kontext.
In comparison, the UMs, particularly the open-source ones, can suffer from unsatisfactory image quality, as verified by results in 
Figure~\ref{fig:image quality Qualitative results}.
This aligns with the gap between open-source UMs and the \emph{editing-based} approach on the image perception quality metric in Table \ref{tab:editing-based} and Table \ref{tab:UM-based}.

We provide a more detailed qualitative analysis of the WCA and WDQ metrics in Appendix \ref{app:high-scoring and low-scoring}.

\subsection{Evaluation of Metric Effectiveness and Robustness}

\label{sec:human evaluation}

\begin{table}[t]
\centering
\caption{Correlation between our metrics and human evaluations. Inter-human agreement is included for comparison.}
\label{tab:correlation_results}
\renewcommand{\arraystretch}{1.2} 
\resizebox{\columnwidth}{!}{%
  \setlength{\tabcolsep}{2pt} 
  \begin{tabular}{l *{6}{c}}

  \toprule
  \multirow{2}{*}{Metric} & \multicolumn{3}{c}{Human-Metric} & \multicolumn{3}{c}{Human-Human} \\
  \cmidrule(lr){2-4} \cmidrule(lr){5-7}
  & Pearson & Spearman & Kendall & Pearson & Spearman & Kendall \\
  \midrule
  VCIF & 0.81 & 0.79 & 0.66 & 0.83 & 0.84 & 0.74 \\
  WDQ & 0.72 & 0.74 & 0.61 & 0.73 & 0.75 & 0.65 \\
  WCA & 0.88 & 0.86 & 0.74 & 0.76 & 0.77 & 0.68 \\
  WIF & 0.76 & 0.75 & 0.73 & 0.75 & 0.74 & 0.69 \\
  \bottomrule
  \end{tabular}
} 
\setlength{\tabcolsep}{6pt}
\renewcommand{\arraystretch}{1.0}

\end{table}

To validate the effectiveness of the proposed metrics, we evaluate their correlation with evaluations from 20 human experts for website design, with each expert evaluating 100 samples. We bypass the IPQ metric because it exactly follows prior works ~\cite{ku2023viescore}. 
We calculate the Pearson, Spearman, and Kendall correlation coefficients and calculate the inter-human correlation as a reference.  
As shown in Table \ref{tab:correlation_results}, the human-metric correlation for both visual content instruction following and webpage design quality is close to the human-human correlation. The human-metric correlation for webpage content appeal and the human-metric agreement for webpage instruction even surpass the human-human results. This demonstrates that our metrics align well with human evaluations, proving their effectiveness. See Appendix \ref{app:corrlation analysis} for details of human evaluation.

Beyond alignment with human evaluations, we further verify the stability of our LLM-as-a-judge metrics. We conduct three independent evaluation runs on a subset of generation results sampled from various approaches and models, calculating the standard deviation of each metric. As shown in Appendix \ref{app:corrlation analysis}, the standard deviations for all metrics fall within a reasonable range ($\sigma \le 0.72$). These results demonstrate the robustness of our metrics.

\section{Improving BAGEL for ProductWebGen via Fine-tuning}

\begin{table*}[t]
\centering
\caption{Fine-tuning results of BAGEL ~\cite{deng2025emerging} and comparison to other baselines. VCIF and IPQ evaluate visual instruction following and image quality. WIF, WDQ, and WCA evaluate webpage instruction following, design quality, and content appeal.}
\label{tab:finetune result}
\small
\renewcommand{\arraystretch}{1.2} 
\begin{tabularx}{\textwidth}{@{} l l *{5}{>{\centering\arraybackslash}X} @{}}
\toprule
&  & \multicolumn{2}{c}{\textbf{Image-related}} & \multicolumn{3}{c}{\textbf{Webpage-related}} \\
\cmidrule(lr){3-4} \cmidrule(lr){5-7}
& & \textbf{VCIF} & \textbf{IPQ} & \textbf{WIF} & \textbf{WDQ} & \textbf{WCA} \\[-1pt]
& & (0-10) & (0-10) & (0-1) & (0-10) & (0-10) \\
\midrule
\multirow{3}{*}{\emph{UM-based}} & BAGEL & 5.84 & 5.43 & 0.40 & 7.26 & 5.61 \\
 & BAGEL-finetuned & 7.14\makebox[0pt][l]{\fontsize{5}{6}\selectfont\hspace{0.2em}\textcolor{red}{(+1.30)}} & 5.97\makebox[0pt][l]{\fontsize{5}{6}\selectfont\hspace{0.2em}\textcolor{red}{(+0.54)}} & 0.66\makebox[0pt][l]{\fontsize{5}{6}\selectfont\hspace{0.2em}\textcolor{red}{(+0.26)}} & \textbf{8.06}\makebox[0pt][l]{\fontsize{5}{6}\selectfont\hspace{0.2em}\textcolor{red}{(+0.80)}} & \textbf{7.97}\makebox[0pt][l]{\fontsize{5}{6}\selectfont\hspace{0.2em}\textcolor{red}{(+2.36)}} \\
& Gemini-2.5-Flash-Image  & \textbf{8.15} & \textbf{8.35} & 0.84 & 7.92 & 7.31 \\
\midrule
\multicolumn{2}{c}{\emph{Editing-based} (the best one)} & 6.45 &8.00  & \textbf{0.87} & 7.98 & 7.45 \\
\bottomrule
\end{tabularx}
\renewcommand{\arraystretch}{1.0} 
\end{table*}

To narrow the gap between open-source UMs and Gemini-2.5-Flash-Image for multimodal webpage generation, we construct a training dataset containing 1k samples, dubbed ProductWebGen-1k. 
We fine-tune the open-source UM BAGEL ~\cite{deng2025emerging} on it in this section.

\textbf{Dataset Curation}
According to the task configuration, each training sample consists of three components: a user instruction, a group of five product images (to be displayed on the webpage), and the HTML code. 
For the product images, 
we collect 2,000 groups of product images from the internet and, through filtering, obtain a final set of 1,000 groups.
After filtering, the images in each group satisfy one of the four aforementioned consistency categories for defining visual content instruction. 
For HTML synthesis, we use GPT-4o to provide a basic draft and use Gemini-2.5-Flash to refine it into a high-quality final version. 
The visual content and webpage instructions are both constructed with the aid of (multimodal) LLMs. See Appendix \ref{app:finetune dataset construction} for more detailed process.

\textbf{Training Details}
We fine-tune BAGEL for 6 epochs, with a learning rate of 2.5e-5 and a batch size of 8, on 8 GPUs. We jointly train with the cross-entropy loss $\mathcal{L}_{\text{CE}}$ for HTML generation and the mean-squared error $\mathcal{L}_{\text{MSE}}$ for image diffusion: $\mathcal{L}_{\text{Total}} =  \mathcal{L}_{\text{CE}} + \lambda \mathcal{L}_{\text{MSE}}$, where $\lambda$ is a trade-off factor. 
We ablate this trade-off factor in Table~\ref{tab:lambda_ablation}. Setting $\lambda=8$ attains the highest VCIF and IPQ scores, but reduces all webpage-related metrics relative to $\lambda=4$. Conversely, $\lambda=1$ yields the highest WIF score while producing substantially weaker image-related results. We therefore use $\lambda=4$, which provides a balanced configuration and achieves the best WDQ and WCA scores.
As discussed in Section \ref{sec:Automatic Evaluation}, verbose HTML code can hinder image generation, so we opt for a training policy aligned with the \emph{UM-based} approach instead of the \emph{UM-based (HTML)} one. We name the resultant model \emph{BAGEL-finetuned}.

\begin{table}[t]
\centering
\caption{Ablation of the loss trade-off factor $\lambda$ for fine-tuning BAGEL.}
\label{tab:lambda_ablation}
\small
\setlength{\tabcolsep}{3pt}
\begin{tabularx}{\columnwidth}{@{} *{6}{>{\centering\arraybackslash}X} @{}}
\toprule
$\lambda$ & \textbf{VCIF} & \textbf{IPQ} & \textbf{WIF} & \textbf{WDQ} & \textbf{WCA} \\
\midrule
$1$ & $6.52$ & $5.27$ & $\mathbf{0.68}$ & $7.94$ & $7.75$ \\
$4$ & $7.14$ & $5.97$ & $0.66$ & $\mathbf{8.06}$ & $\mathbf{7.97}$ \\
$8$ & $\mathbf{7.18}$ & $\mathbf{6.04}$ & $0.57$ & $7.65$ & $7.63$ \\
\bottomrule
\end{tabularx}
\end{table}

\label{sec:4}

\textbf{Results} We employ the \emph{UM-based} approach for evaluation, with results summarized in Table \ref{tab:finetune result}.
As shown, BAGEL-finetuned improves over BAGEL across all five metrics.
Specifically, fine-tuning increases the visual content instruction following score from 5.84 to 7.14 and the image perception quality score from 5.43 to 5.97, while improving webpage instruction following from 0.40 to 0.66, webpage design quality from 7.26 to 8.06, and webpage content appeal from 5.61 to 7.97. BAGEL-finetuned surpasses the strongest \emph{editing-based} baseline in visual content instruction following (7.14 vs.\ 6.45), and achieves the best webpage design quality and webpage content appeal scores among all compared approaches. These consistent gains demonstrate the effectiveness of ProductWebGen-1k for improving both image generation and webpage generation capabilities of BAGEL. Qualitative comparisons in Appendix \ref{app:bagel fintune visual comparison} further illustrate the improvements in webpage quality and visual consistency.

\section{Related Work}

\textbf{Webpage Generation} Existing webpage generation benchmarks predominantly study the conversion of visual designs into front-end code: Sketch2Code ~\cite{robinson2019sketch2code} uses sketches, while Pix2Code ~\cite{beltramelli2018pix2code} and Design2Code ~\cite{si2024design2code} focus on screenshots or rendered webpages. Recent MLLM-based methods further advance visually conditioned webpage code generation ~\cite{wu2025mllm,gui2025uicopilot,wan2025divide}. In parallel, datasets such as WebSight ~\cite{laurenccon2024unlocking} and WebCode2M ~\cite{gui2025webcode2m} provide increasingly large-scale training resources. Some benchmarks further extend the evaluation scope: MRWeb~\cite{wan2024mrweb} examines multi-page generation, while Interactive2Code~\cite{xiao2024interaction2code} focuses on interactive elements. 
Although most of these benchmarks are multimodal, the multimodality is primarily on the input side (e.g., screenshots or sketches), while the output side remains webpage code only. ProductWebGen addresses a different, complementary challenge: given a source product image together with visual content and webpage instructions, a model must jointly generate renderable HTML and multiple product images that remain grounded in the source image and consistent with the visual instruction. Accordingly, our benchmark evaluates image-related properties, including visual content instruction following and image perception quality, in addition to webpage generation quality.

\textbf{Unified Multimodal Model}
Recently, many studies have explored unified models for both image understanding and generation ~\cite{ma2025janusflow,liao2025mogao,zhou2025transfusion,lin2025toklip,wu2024liquid}. 
Some works, such as Chameleon ~\cite{team2024chameleon} and EMU3 ~\cite{wang2024emu3}, adopt a unified token space to process interleaved image--text sequences. 
Others focus on reducing information loss or enhancing capacity: Orthus ~\cite{kou2024orthus} uses modality-specific heads for text and image, while BAGEL ~\cite{deng2025emerging} employs a Mixture-of-Transformer-Experts design. 
Show-o2 ~\cite{xie2025show} combines autoregressive modeling with flow matching for text and visual generation.  
Ovis-U1 ~\cite{wang2025ovis} introduces a multi-stage training framework with a novel visual decoder, while OmniGen2 ~\cite{wu2025omnigen2} separates text and image generation to avoid suboptimal parameter sharing. 
In this work, we fine-tune BAGEL on our curated webpage generation dataset and demonstrate that unified models can generate multiple consistent images.

\section{Conclusion}

In this paper, we introduce ProductWebGen, a novel benchmark designed to systematically evaluate the capacity of multimodal generative models for multimodal webpage generation. It requires models to jointly generate renderable HTML code and visually consistent images in response to complex, mixed-modality instructions. We design and systematically evaluate two novel evaluation workflows, finding that the \emph{editing-based} approach overall excels at webpage instruction following, design quality, and content appeal, while the \emph{UM-based} approach shows a distinct advantage in maintaining visual content consistency. Our results also highlight a significant performance gap between open-source unified models and the closed-source Gemini-2.5-Flash-Image. To bridge this gap, we construct a training dataset, ProductWebGen-1k. By fine-tuning the open-source UM BAGEL, we show consistent improvements across metrics, validating our dataset's effectiveness and significantly narrowing the capability gap.

\begin{acks}
This work was supported by Shanghai Key Technology R\&D Program ``New Generation of Information Technology'' (No. 25511103700), NSF of China (Nos. 62306176, 92470118), CCF-ALIMAMA TECH Kangaroo Fund (NO. CCF-ALIMAMA OF 2025010), Kuaishou Technology, and Ant Group. 
\end{acks}

\clearpage

\bibliographystyle{ACM-Reference-Format}
\bibliography{sample-base}

\appendix

\clearpage


\section{Prompts for Data Curation}

\label{app:visual content instruction generation prompt}

The complete system prompt in the user instruction is shown in Figure \ref{fig:system prompt}.
There are four types of visual content instructions, and the prompts for generating each type of instruction are shown in Figure \ref{fig:background}, Figure \ref{fig:watermark}, Figure \ref{fig:character}, and Figure \ref{fig:angle}. The prompt for extracting webpage instructions from the synthesized HTML code is presented in Figure \ref{fig:webpage instruction}.

\section{LLM-as-a-judge Prompt}

\label{app:metric prompt}

The prompts for the four newly proposed metrics: visual content instruction following, webpage instruction following, webpage design quality, and webpage content appeal, are shown in Figure \ref{vcif-background}--\ref{vcif-perspective}, Figure \ref{wif}, Figure \ref{wdq}, and Figure \ref{wca} respectively. For the visual content instruction following metric, we design specialized prompts for each type of instruction.

\section{Experimental Setup Details}
\label{app:experiment detail}

For the LLMs employed in the \emph{editing-based} approach and the LLM-as-a-judge evaluation, we utilize the API provided by the OpenRouter platform. We adopt the default parameter settings provided by the OpenRouter platform. We provide the parameter settings for the unified models and image editing models in Table \ref{tab:param}.

\begin{table*}[t]
\centering
\caption{Category-level performance margins for three key conclusions. ``Gemini-Image'' denotes Gemini-2.5-Flash-Image, and ``Editing-based'' denotes Gemini-2.5-Flash + Qwen-Image-Edit. Positive values indicate that the first method in each column outperforms the second.}
\label{tab:category_generalization}
\scriptsize
\setlength{\tabcolsep}{3pt}
\resizebox{\textwidth}{!}{%
\begin{tabular}{@{} l ccc ccc @{}}
\toprule
\textbf{Category} & \textbf{\makecell{VCIF\\(Gemini-Image $-$ Editing-based)}} & \textbf{\makecell{VCIF\\(Gemini-Image $-$ BAGEL)}} & \textbf{\makecell{IPQ\\(Gemini-Image $-$ BAGEL)}} & \textbf{\makecell{WIF\\(Editing-based $-$ BAGEL)}} & \textbf{\makecell{WDQ\\(Editing-based $-$ BAGEL)}} & \textbf{\makecell{WCA\\(Editing-based $-$ BAGEL)}} \\
\midrule
Food & 1.34 & 2.15 & 1.83 & 0.43 & 0.64 & 2.24 \\
Apparel & 2.19 & 0.35 & 3.52 & 0.48 & 0.54 & 1.64 \\
Beauty & 0.28 & 2.40 & 3.16 & 0.42 & 0.70 & 2.25 \\
Household supplies & 0.52 & 3.86 & 2.44 & 0.52 & 0.60 & 1.85 \\
Digital products & 1.75 & 2.28 & 3.78 & 0.46 & 1.20 & 2.42 \\
Appliances & 1.32 & 2.42 & 2.32 & 0.38 & 0.54 & 2.34 \\
Baby products & 2.25 & 2.34 & 0.92 & 0.49 & 1.14 & 2.73 \\
Office supplies & 2.07 & 3.44 & 2.79 & 0.56 & 0.78 & 1.61 \\
Pet supplies & 1.76 & 3.74 & 2.74 & 0.43 & 0.92 & 2.03 \\
Furniture & 0.90 & 1.61 & 1.06 & 0.51 & 0.74 & 2.07 \\
Sports & 0.57 & 1.33 & 3.59 & 0.45 & 1.02 & 1.63 \\
Jewelry & 1.96 & 4.32 & 2.86 & 0.42 & 1.00 & 1.90 \\
Kitchenware & 1.85 & 3.33 & 2.11 & 0.52 & 0.72 & 2.44 \\
\bottomrule
\end{tabular}%
}
\end{table*}

\begin{table*}[t]
    \centering
    \caption{Parameter settings for the unified models and image editing models. The parameter names align with the official code implementations.}
    \label{tab:param}
    \resizebox{\textwidth}{!}{
        \begin{tabular}{lllll}
            \toprule
            \textbf{BAGEL} & \textbf{Ovis-U1} & \textbf{OmniGen2} & \textbf{Qwen-Image-Edit} & \textbf{FLUX.1-Kontext} \\
            \midrule
            \makecell[tl]{
                seed=42 \\
                do\_sample=False \\
                num\_timesteps=50 \\
                cfg\_text\_scale=5.0 \\
                cfg\_img\_scale=1.5 \\
                cfg\_interval=[0.0, 1.0] \\
                timestep\_shift=3.0 \\
                cfg\_renorm\_min=0.0 \\
                cfg\_renorm\_type=``text\_channel''
            } & 
            \makecell[tl]{
                seed=42 \\
                do\_sample=False \\
                steps=50 \\
                txt\_cfg=7.5 \\
                img\_cfg=1.5
            } & 
            \makecell[tl]{
                seed=0 \\
                do\_sample=False \\
                num\_inference\_step=50 \\
                text\_guidance\_scale=5.0 \\
                image\_guidance\_scale=1.5 \\
                cfg\_range\_start=0.0 \\
                cfg\_range\_end=1.0
            } & 
            \makecell[tl]{
                seed=0 \\
                num\_inference\_steps=50 \\
                true\_cfg\_scale=4.0
            } & 
            \makecell[tl]{
                seed=0 \\
                num\_inference\_steps=28 \\
                guidance\_scale=2.5
            } \\
            \bottomrule
        \end{tabular}
    }
\end{table*}

\section{Human Evaluation Details}
\label{app:corrlation analysis}

\begin{figure*}[t]
    \centering
    \includegraphics[width=1\linewidth]{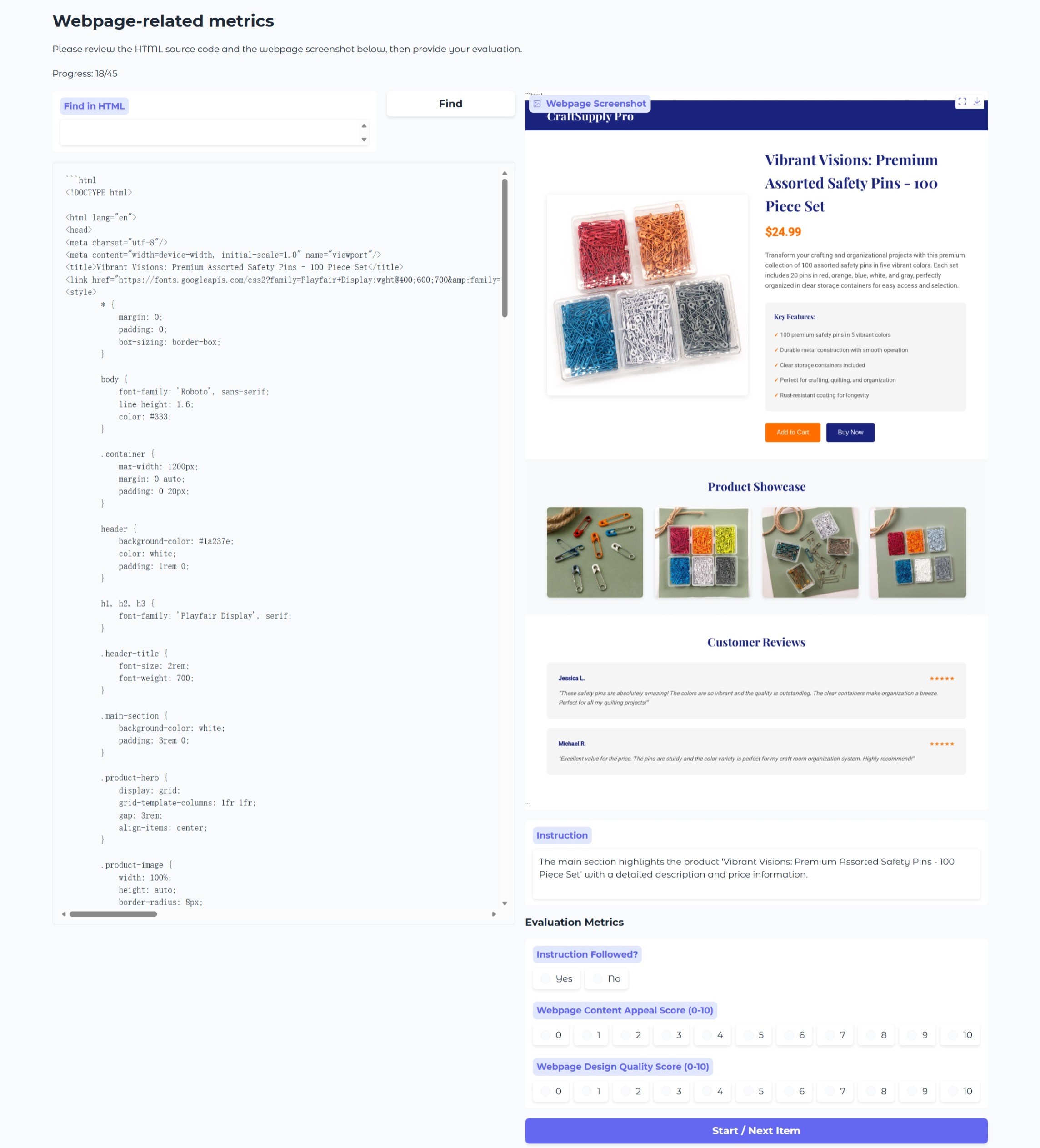}
    \caption{An example of the scoring interface. Our interface is easy to use. It includes clear textual instructions, supports zooming images and searching within the code, and can also record the experts’ evaluation results.}
    \label{fig:scoring}
\end{figure*}

We recruit 20 experts in website design and provide them with detailed evaluation guidelines to help them score the webpages and images generated by the models. For each metric, we select 100 samples using a stratified sampling strategy to ensure a balanced distribution across different approaches and models. These sample sets are mutually exclusive across metrics. An example of the scoring interface is shown in Figure \ref{fig:scoring}. After collecting the experts' evaluations, we organize the results and calculate the correlations, which demonstrates the effectiveness of the metrics we propose.

We conduct three independent evaluation runs on a subset of generation results sampled from various approaches and models, calculating the standard deviation of each metric. The results are presented in Table \ref{tab:robustness}.

\begin{table}[t]
\centering
\caption{Standard deviation of our metrics calculated over three independent evaluation runs.}
\label{tab:robustness}
\begin{tabular}{lccccc}
\toprule
\textbf{Metric} & \textbf{VCIF} & \textbf{IPQ} & \textbf{WIF} & \textbf{WDQ} & \textbf{WCA} \\
\midrule
\textbf{Std ($\sigma$)} & 0.72 & 0.32 & 0.05 & 0.20 & 0.19 \\
\bottomrule
\end{tabular}
\end{table}

\section{Case Studies on WCA and WDQ Metrics}
\label{app:high-scoring and low-scoring}
\begin{figure*}[t]
    \centering
    \includegraphics[width=1\linewidth]{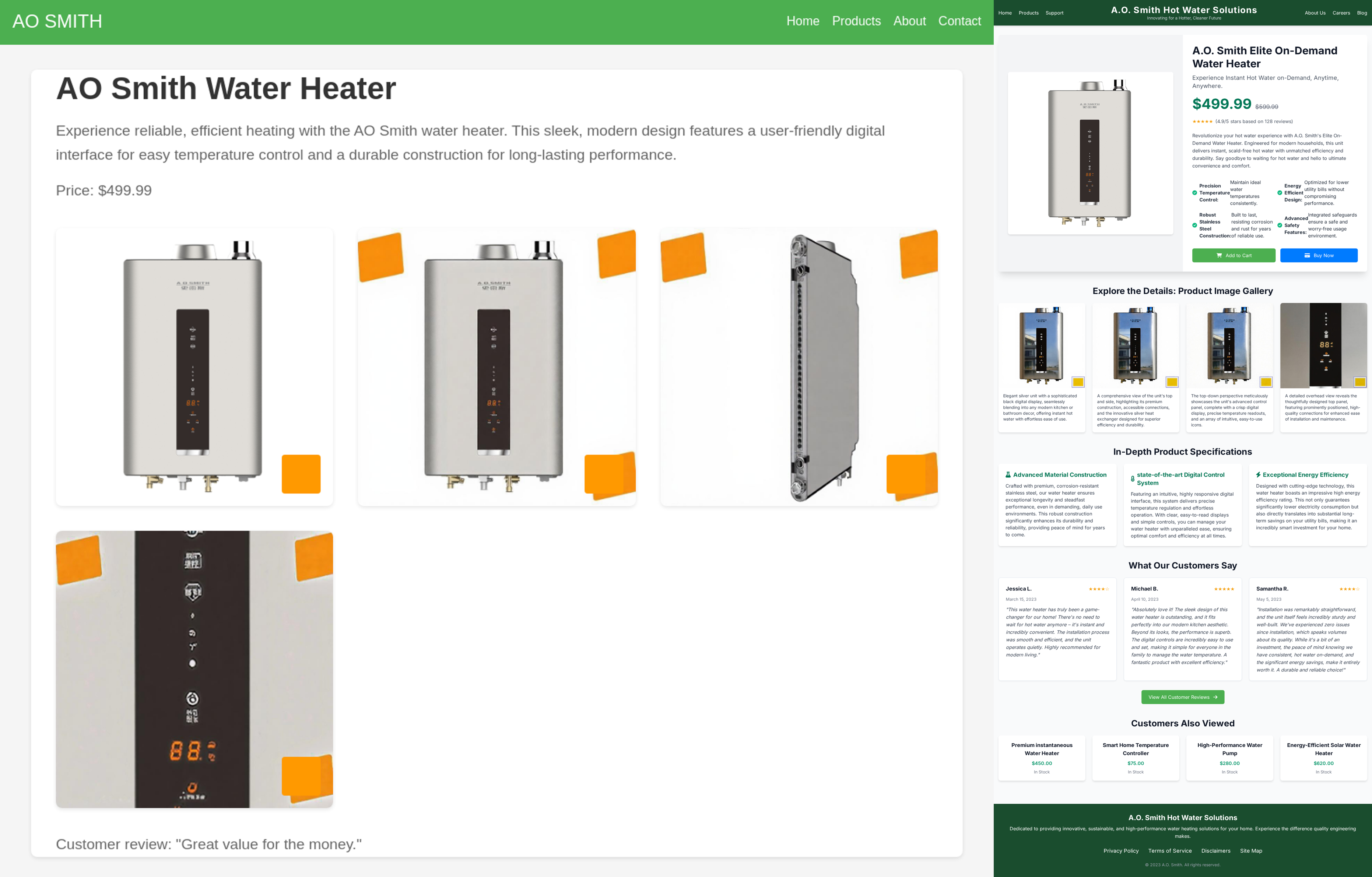}
    \caption{Comparison of low and high scoring samples for Webpage Content Appeal (WCA)}
    \label{fig:WCA}
\end{figure*}

Figure \ref{fig:WCA} shows webpages generated by different models on the same test data. The Webpage Content Appeal (WCA) score of the left webpage is 4, while that of the right webpage is 8.
The webpage on the right effectively drives purchase intent by featuring a rich array of conversion-focused components, such as a structured `In-Depth Product Specifications' section with visual icons, a dedicated `What Our Customers Say' block displaying specific star ratings and testimonials, and prominent `Add to Cart' and `Buy Now' buttons. Conversely, the webpage on the left presents only a rudimentary product description and a single line of unformatted review text, devoid of these critical persuasive elements. This comparison demonstrates that the WCA metric effectively captures the disparity in content appeal and accurately reflects the webpage's capability to drive customer interest.

\begin{figure*}[t]
    \centering
    \includegraphics[width=1\linewidth]{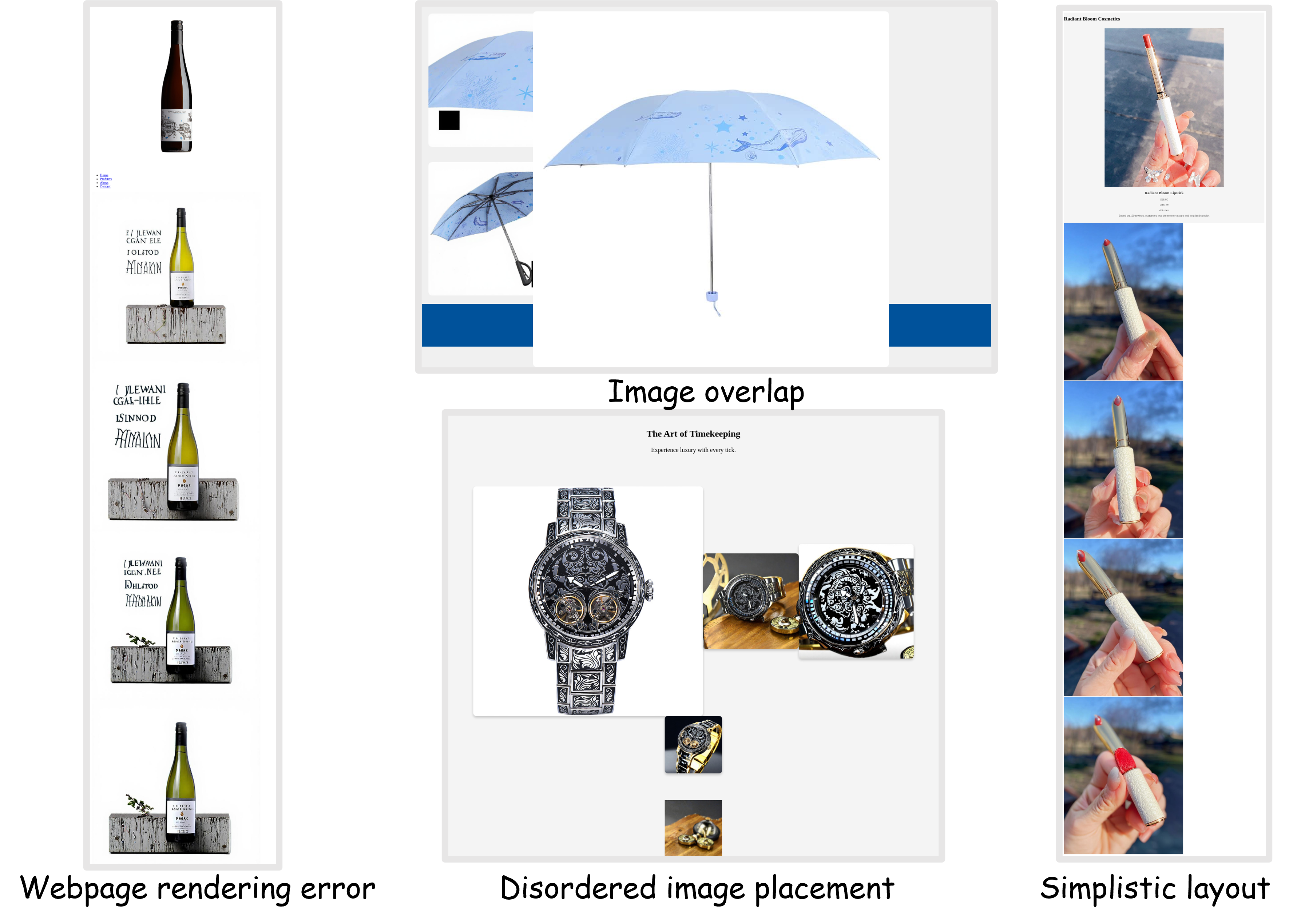}
    \caption{Examples of generated webpages with low Webpage Design Quality (WDQ) scores.}
    \label{fig:WDQ}
\end{figure*}

As shown in Figure \ref{fig:WDQ}, we observe several common reasons for lower Webpage Design Quality (WDQ) scores, including webpage rendering error, disordered image placement, image overlap, and simplistic layout, all of which degrade the webpage design quality.

\section{Construction Details of ProductWebGen-1k}

\label{app:finetune dataset construction}

Here, we provide a detailed description of the construction process of the ProductWebGen-1k fine-tuning dataset. Overall, the dataset is built through the following five steps:

\begin{enumerate}[left=0pt]
    \item \textbf{Product Images Collection and Preliminary Filtering: }We collect a large number of product display images from popular e-commerce websites, corresponding to the product categories in the benchmark. For each product, we collect five display images. First, we apply visual-quality screening: PaddleOCR is used to detect and remove images containing advertising text, and we further exclude images with watermarks, low resolution, or excessive blur. This produces 2,000 candidate groups of product images. Then, we use Qwen2.5-VL-32B-Instruct to select a suitable source image for each product.

    \item \textbf{Further Fine-Grained Filtering: }To improve the model's ability to follow visual instructions, we require the product images in the fine-tuning dataset to satisfy one of four types of visual content instructions. Therefore, we meticulously craft filtering prompts and use Qwen2.5-VL-32B-Instruct to filter product images that meet the criteria for each instruction type. Due to the scarcity of data satisfying the ``ensuring coherent perspectives'' criterion, we leverage the Amazon Berkeley Objects dataset  ~\cite{collins2022abo}. More than 8,200 products in this dataset include a sequence of 72 images, capturing the product every 5º in azimuth. We select five images with continuously changing perspectives for each product. Finally, we obtain 1,000 groups of product images, distributed as follows: using the same human model (140), ensuring coherent perspectives (260), maintaining a consistent background (300), and applying an identical watermark (300).

    \item \textbf{Visual Content Instructions and Text Descriptions Generation: }We utilize GPT-4o to write a suitable visual content instruction for each group of filtered product images. Next, we prompt Gemini-2.5-Flash to generate detailed descriptions for the images except the source one, based on each group of images and the visual content instruction.

    \item \textbf{HTML Code Generation: }We employ a ``draft-then-refine'' method to synthesize high-quality HTML code with LLMs. Specifically, we first prompt the cost-effective GPT-4o to generate a basic webpage based on each group of product images and their text descriptions. Then, we use the more powerful Gemini-2.5-Flash to refine the simple HTML code, producing a high-quality final version.

    \item \textbf{Final User Instruction Generation: }Following the same methodology as adopted in the benchmark construction, we use GPT-4o to generate webpage instructions based on the HTML code. This is then combined with the system prompt and the visual content instruction to create the final user instruction.
    
\end{enumerate}

Through the systematic process, we curate a high-quality fine-tuning dataset that integrates product images, user instructions, and corresponding webpage HTML code. 

\section{Qualitative Comparison between BAGEL and BAGEL-finetuned}
\label{app:bagel fintune visual comparison}

\begin{figure*}[t]
    \centering
    \includegraphics[width=\linewidth, height=0.95\textheight, keepaspectratio]{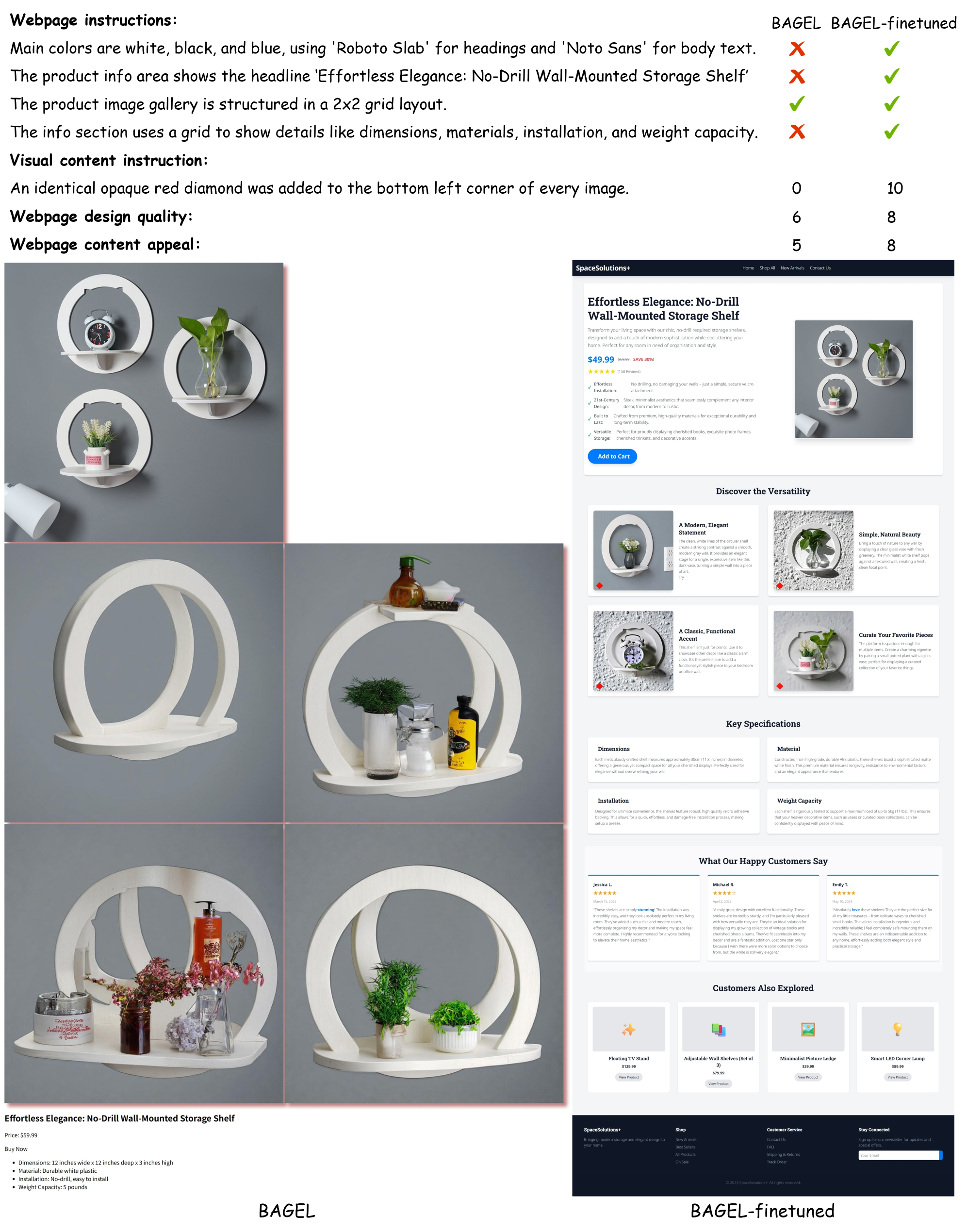}
    \caption{Qualitative comparison between the original BAGEL (left) and BAGEL-finetuned (right).}
    \label{fig:bagel_watermark}
\end{figure*}

\begin{figure*}[t]
    \centering
    \includegraphics[width=\linewidth, height=0.95\textheight, keepaspectratio]{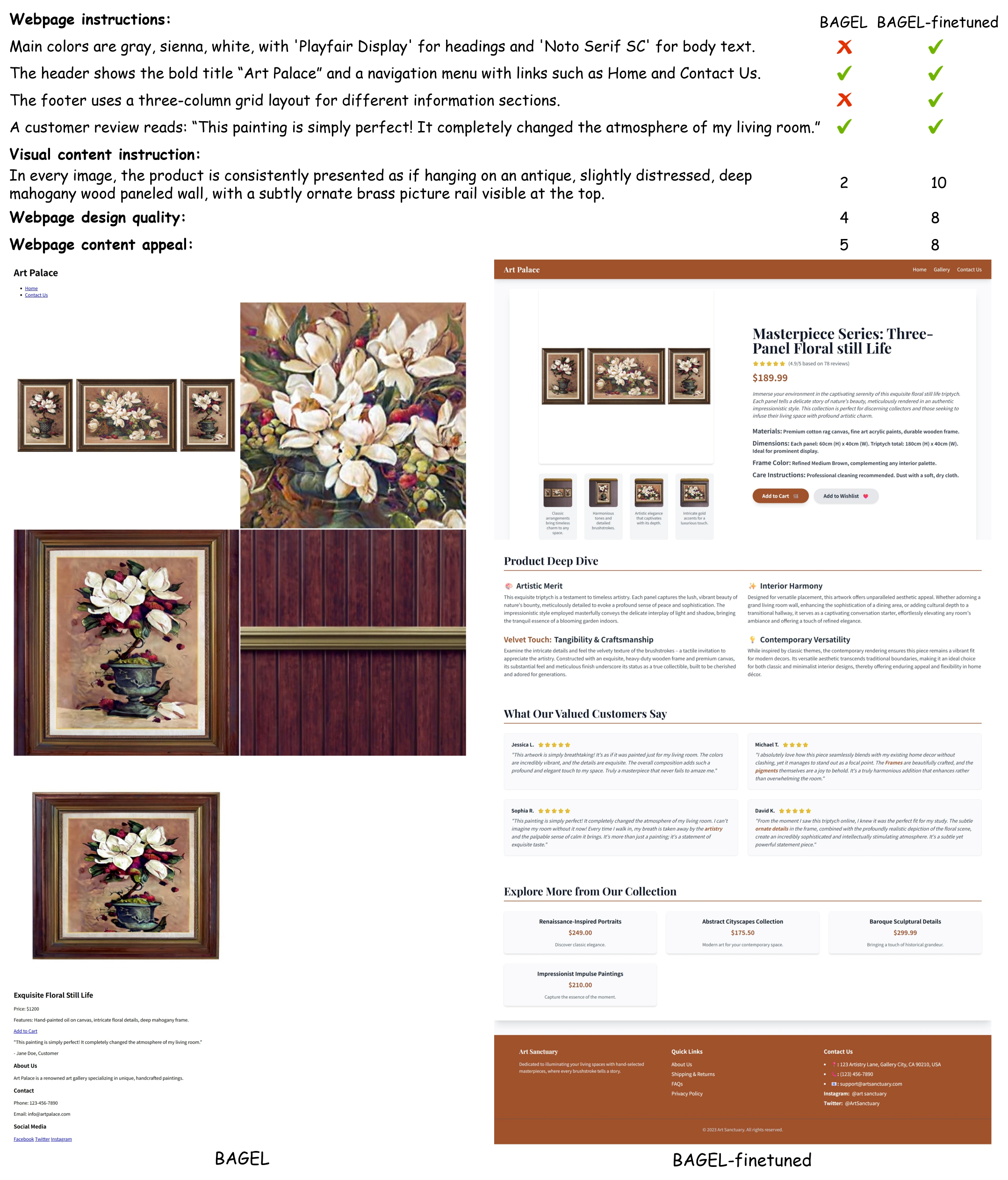}
    \caption{Qualitative comparison between the original BAGEL (left) and BAGEL-finetuned (right).}
    \label{fig:bagel_background}
\end{figure*}

\begin{figure*}[t]
    \centering
    \includegraphics[width=\linewidth, height=0.95\textheight, keepaspectratio]{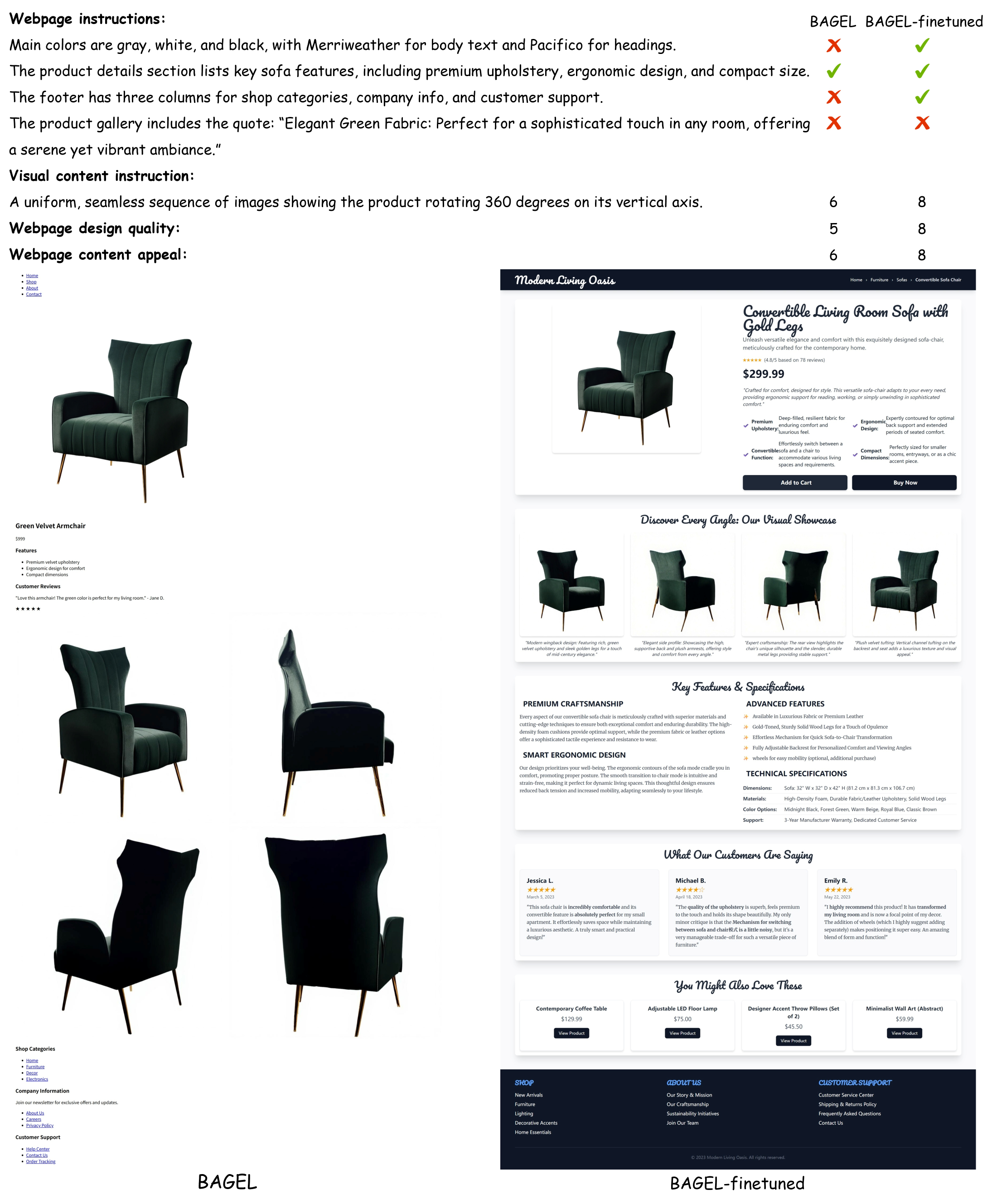}
    \caption{Qualitative comparison between the original BAGEL (left) and BAGEL-finetuned (right).}
    \label{fig:bagel_viewpoint}
\end{figure*}

\begin{figure*}[t]
    \centering
    \includegraphics[width=\linewidth, height=0.95\textheight, keepaspectratio]{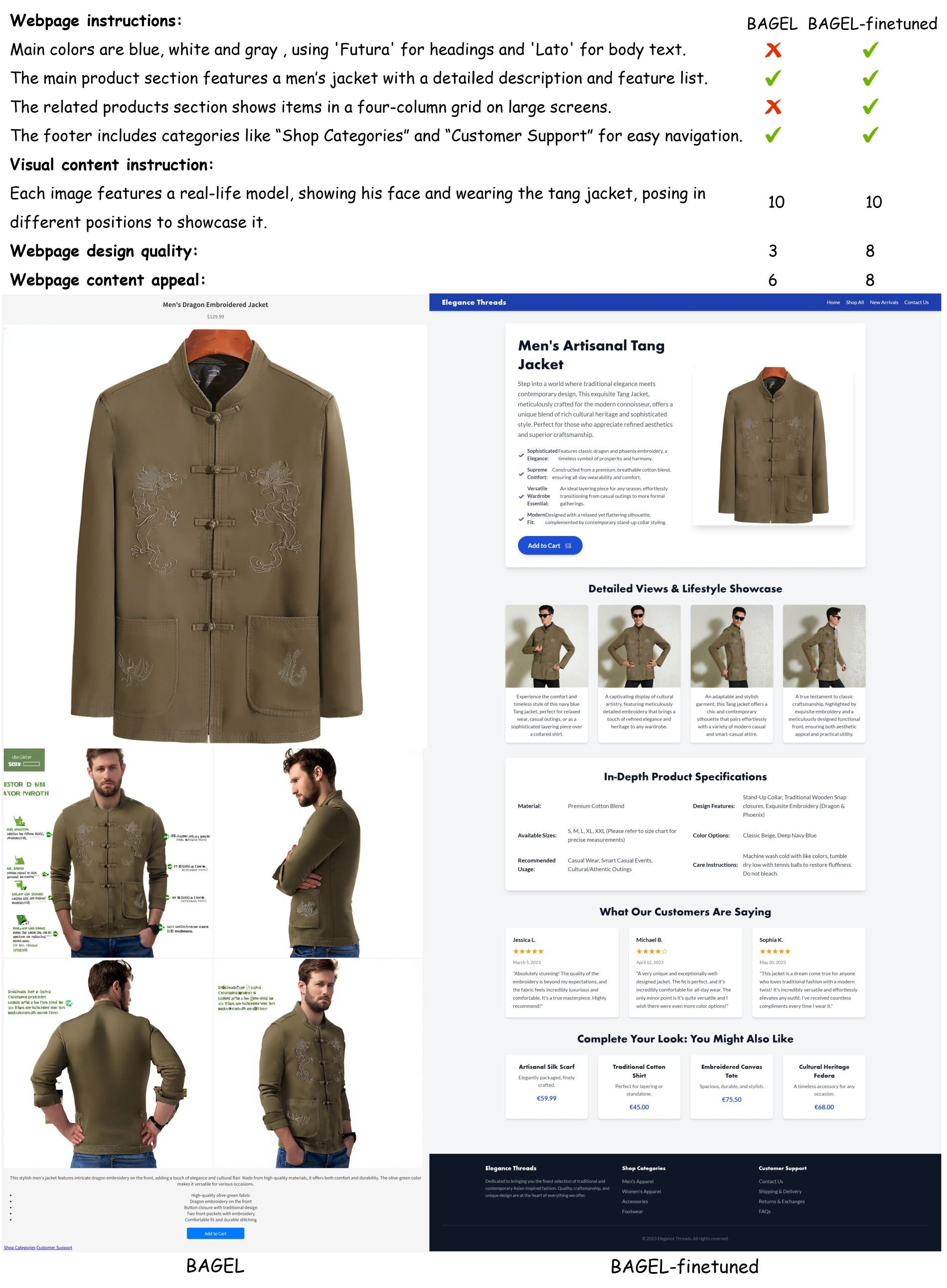}
    \caption{Qualitative comparison between the original BAGEL (left) and BAGEL-finetuned (right).}
    \label{fig:bagel_model}
\end{figure*}

We present a comprehensive qualitative comparison between BAGEL and BAGEL-finetuned in Figure \ref{fig:bagel_watermark}, Figure \ref{fig:bagel_background}, Figure \ref{fig:bagel_viewpoint} and Figure \ref{fig:bagel_model}. As illustrated in these examples, the original BAGEL often struggles with generating renderable HTML code, resulting in disordered layouts, unreasonable image sizes, and a lack of aesthetic appeal. Furthermore, it frequently fails to adhere to the visual content instructions. In contrast, BAGEL-finetuned exhibits a substantial improvement. It generates structurally valid and visually appealing webpages that strictly follow layout constraints (e.g., grid layout, font specifications). The generated webpages are also enriched with detailed content and components that attract customers, such as user reviews, discount information, and prominent ``add-to-cart'' buttons. Simultaneously, it produces product images that faithfully follow the visual content instructions. These visually distinct improvements confirm the quantitative gains in metrics reported in Section \ref{sec:4}.

\begin{figure*}[t]
    \centering       
    \begin{tcolorbox}[
        title=System prompt in the user instruction,
        fonttitle=\bfseries,
        colback=blue!5!white,
        colframe=blue!75!black,
        boxrule=1pt
    ]
    \begin{lstlisting}[style=promptstyle]
You are provided a product image. Generate a complete, single-file HTML page that showcases the product. The page MUST contain exactly five <img> tags in total: one for the main product image and four for additional product showcase images. {Visual Content Instruction} No other <img> tags are permitted on the page. Use the provided image as the page's main product image. Every <img> tag MUST include an exceptionally clear, detailed, and descriptive `alt` attribute. The `alt` text must comprehensively describe the image content, including the product's appearance, texture, color, angle, and any relevant context or surrounding elements. You may include elements such as a product title and price, a detailed product description, features, selectors (color/size), CTAs (Add to cart / Buy now), a review section, or any other sections you find useful. Organize and combine elements as you see fit to make an effective product page. {Webpage Instruction} Output only the complete HTML code (no extra explanation or comments outside the HTML).
    \end{lstlisting}
    \end{tcolorbox}
    
    \caption{System prompt in the user instruction.} 
    \label{fig:system prompt}   
\end{figure*}

\begin{figure*}[t] 
    \centering

    \begin{tcolorbox}[
        title=Prompt for generating background consistency visual content instruction,
        fonttitle=\bfseries,
        colback=blue!5!white,
        colframe=blue!75!black,
        boxrule=1pt
    ]
    \begin{lstlisting}[style=promptstyle]
# Role
Act as a highly creative AI Prompt Engineer and E-commerce Art Director. Your sole task is to invent and describe a consistent, visually rich background or scene for a series of product images.
# Context
I will provide a main product image. Based on this product, you must invent a compelling and highly consistent visual setting for it. This setting must be identical across a whole series of secondary images.
# Prohibited Content
1. Lighting: Avoid any mention of lighting, shadows, or illumination.
2. Abstract Meaning: Do not explain the purpose or mood. Focus only on the visual description.
# Core Task
Your entire focus is on the environment around the product. This can be a simple surface, a recurring prop, or a full lifestyle scene. You must be highly creative and specific, moving beyond simple colored backgrounds.
# Instruction Crafting Rules
1. Be Specific: Describe the surface, props, colors, and composition in detail.
2. Be AI-Friendly: Phrase the instruction as a clear description of the final images.
3. Emphasize Consistency: Use wording like "across all images," "for each image," "the exact same."
# Excellent Examples of Final Instructions
* "Across all images, the identical product is presented on a rough, dark slate stone surface, consistently surrounded by a few scattered fresh green moss elements."
* "For each image, the product is placed on the exact same clean, light-grained oak wood desk, next to a recurring, out-of-focus small succulent in a white ceramic pot."
* "A series of images where the product is consistently positioned on the bottom-right corner of a uniform, textured, beige linen fabric background."
* "In every image, the product rests on the same single, suspended, polished concrete slab against an otherwise empty, neutral gray background."
# Output Format
Your output must be a single line, containing only the instruction.
Instruction: [Your specific, consistent-background, single-sentence instruction here]
Now, analyze the main product image I provide and generate the consistent background/scene instruction.
    \end{lstlisting}
    \end{tcolorbox}
    \caption{Prompt for generating background consistency visual content instruction.}
    \label{fig:background}
    \end{figure*}
    
\begin{figure*}[t] 
    \centering
\begin{tcolorbox}[
        title=Prompt for generating watermark consistency visual content instruction,
        fonttitle=\bfseries,
        colback=blue!5!white,
        colframe=blue!75!black,
        boxrule=1pt
    ]
    \begin{lstlisting}[style=promptstyle]
# Role
Act as an AI Prompt Engineer and Graphic Designer. Your sole task is to generate a meta-prompt that describes a series of product images consistently featuring a graphic overlay in one of the four corners.
# Context
I will provide a main product image. Your task is to generate an instruction for a series of images, where the single defining feature is a consistent graphic element (a shape, a block of color, an icon) placed in the exact same corner position in every image.
# Prohibited Content
1.  Lighting: Avoid any mention of lighting, shadows, or illumination.
2.  Abstract Meaning: Do not explain the purpose or mood. Focus only on the visual description of the graphic element.
# Core Task
Your entire focus is to define the recurring graphic element. You must specify its shape, color, and its position, which must be strictly one of the four corners (top-left, top-right, bottom-left, bottom-right).
# Instruction Crafting Rules
1.  **Be Specific:** Clearly state the shape, color (using natural language), and the precise corner location.
2.  **Be AI-Friendly:** Phrase the instruction as a clear description of the recurring graphic element in a series of images.
3.  **Emphasize Consistency:** Use explicit wording like "in every image," "a recurring," "consistently placed in the," "identical."
# Excellent Examples of Final Instructions
* "For each image in the series, a recurring solid red rectangular block is consistently present in the bottom-left corner."
* "A series of images where every image features the identical semi-transparent, soft gray circle in the top-right corner."
* "In every image of the series, a recurring simple, white leaf icon is consistently positioned in the bottom-right corner."
* "Across all images, an identical solid black square is consistently placed in the top-left corner."
# Output Format
Your output must be a single line, containing only the instruction.
**Instruction:** [Your specific, graphic-overlay, single-sentence instruction here]
---
Now, analyze the main product image I provide and generate the graphic overlay instruction.    \end{lstlisting}
    \end{tcolorbox}
    \caption{Prompt for generating watermark consistency visual content instruction.}
    \label{fig:watermark}    
\end{figure*}

\begin{figure*}[t] 
    \centering
        \begin{tcolorbox}[
        title=Prompt for generating character consistency visual content instruction,
        fonttitle=\bfseries,
        colback=blue!5!white,
        colframe=blue!75!black,
        boxrule=1pt
    ]
    \begin{lstlisting}[style=promptstyle]
# Role
Act as an expert AI Prompt Engineer for fashion and apparel e-commerce. Your sole task is to generate a simple, global consistency rule for a series of model images.
# Context
I will provide a main product image of an apparel item. Your task is to generate a foundational instruction for a series of images. This instruction's only purpose is to state that the **exact same model** and the **exact same background** must be used in every single image. You should not describe what the model is doing (poses, angles, actions, zoom, etc.).
# Prohibited Content
1.  Lighting: Avoid any mention of lighting, shadows, or illumination.
2.  Abstract Meaning: Do not explain the purpose or mood. Focus only on the visual description of consistency.
3.  Specific Actions/Poses: Do not describe the model's specific poses, angles, actions, or the camera distance.
# Core Task
Your entire focus is to state the two core rules of consistency: the model is identical and the background is identical. You can be creative and specific in describing a suitable background, but the instruction should not mention any other details about the images' content.
# Instruction Crafting Rules
1.  **Be Specific:** Your instruction must explicitly mention that the model is the "exact same" and must describe a specific, consistent background (e.g., 'solid neutral gray,' 'a minimalist room setting').
2.  **Be AI-Friendly:** Phrase the instruction as a clear, simple rule for a series of images.
3.  **Emphasize Consistency:** This is the main point. Use explicit wording like "the exact same photo-realistic model," "in every image," "an identical background," "consistently."
# Excellent Examples of Final Instructions
* "For every image in the series, the exact same photo-realistic model is featured, and each image shares an identical solid, soft gray background."
* "A series of images where the apparel is consistently worn by the identical model against a recurring minimalist, out-of-focus interior room setting."
* "Across all images, the product is showcased by the same photo-realistic model, with every shot taking place against an identical off-white studio background."
* "In each image of the series, the identical model is present, and the background is consistently a clean, uniform beige wall."
# Output Format
Your output must be a single line, containing only the instruction.
**Instruction:** [Your simple, model-and-background-consistency, single-sentence instruction here]
---
Now, analyze the main apparel image I provide and generate the simple consistency instruction for the model and background.
"""
\end{lstlisting}
    \end{tcolorbox}
    \caption{Prompt for generating character consistency visual content instruction.}
    \label{fig:character}
\end{figure*}

\begin{figure*}[t] 
    \centering
        \begin{tcolorbox}[
        title=Prompt for generating perspective coherence visual content instruction,
        fonttitle=\bfseries,
        colback=blue!5!white,
        colframe=blue!75!black,
        boxrule=1pt
    ]
    \begin{lstlisting}[style=promptstyle]
# Role
Act as an expert AI Prompt Engineer specializing in product visualization. Your sole task is to generate a meta-prompt for an AI model to create a uniform and continuous rotational view of a product.
# Context
I will provide a main product image. Based on this image, you will generate a single, specific, and purely visual instruction. This instruction will describe a series of images that, together, form a seamless, uniform rotational sequence of the product. This does not have to be a full 360-degree rotation.
# Prohibited Content
1.  Lighting: Avoid any mention of lighting, shadows, or illumination.
2.  Abstract Meaning: Do not explain the purpose or mood. Focus only on the visual description of the product's movement.
# Core Task
Your entire focus is to describe a rotational or tilting view of the product itself against a simple, non-distracting background. The nature of the background is secondary to the motion.
# Instruction Crafting Rules
1.  **Be Specific:** Clearly describe the type of rotational movement (e.g., turning on its vertical axis, tilting from top to front).
2.  **Be AI-Friendly:** Phrase the instruction as a clear description of the final images' sequence.
3.  **Emphasize Movement:** Use explicit wording like "a series of images," "uniform," "seamless sequence," "incrementally rotated," "smoothly turning."
# Excellent Examples of Final Instructions
* "A seamless sequence of images showing the product smoothly turning from a direct front view to a 90-degree side view."
* "A series of images creating a uniform rotational view of the product on its vertical axis against a simple, neutral background."
* "For each image, the product is incrementally tilted from a top-down view to a front-on view."
* "A continuous sequence of images that shows the product rotating 180 degrees from front to back."
# Output Format
Your output must be a single line, containing only the instruction.
**Instruction:** [Your specific, rotational-view, single-sentence instruction here]
---
Now, analyze the main product image I provide and generate the rotational instruction.
\end{lstlisting}
    \end{tcolorbox}
    \caption{Prompt for generating perspective coherence visual content instruction.}
    \label{fig:angle}
    
\end{figure*}

\begin{figure*}[t]
    \centering
    \begin{tcolorbox}[
        title=Prompt for generating webpage instructions,
        fonttitle=\bfseries,
        colback=blue!5!white,
        colframe=blue!75!black,
        boxrule=1pt
    ]
    \begin{lstlisting}[style=promptstyle]
You will be given an HTML code. Analyze the HTML and produce a single output: a raw list of exactly 13 English sentences (an array of 13 strings). Follow these rules exactly:
1 Output format
- Your final output must be a raw list of strings, provided directly without any surrounding text, formatting, or code blocks. The output should strictly follow the format of ["string 1", "string 2", ...].
- The array must contain exactly 13 elements no more no less
2 Sentence rules
- Each array element must be exactly one clear English sentence ending with a single period
- Each sentence must be written in a direct user oriented instruction or suggestion tone aimed at a web designer or developer
- Each sentence must be self contained and focused on a single major aspect described below
- Do not merge multiple aspects into one sentence
3 Major aspects and exact quantities (Produce the sentences in exactly the following order and do not change their sequence)
A Overall webpage style color palette and fonts: 1 sentence
- This single sentence must explicitly describe the primary color palette using plain color words such as purple white black or green and also specify the font family names used on the page for headings and body text
B Specific region concrete content: 4 sentences each about a different prominent region of the HTML
- Each sentence must provide detailed visible textual content and elements present for that region
C Specific region layout characteristic: 4 sentences each about a different prominent region of the HTML
- Each sentence must describe a precise static layout detail.
- Do not include any mention of responsiveness or adaptation to different screens only static layout details
D Explicit quoted text values that appear verbatim in the HTML: 4 sentences
- Each of these four sentences must contain exactly one distinct quoted text value taken verbatim from the HTML enclosed in double quotes
- The quoted value may be short or long and may include punctuation as it appears in the HTML but each quoted value must be distinct and exactly match the visible text in the HTML
4 Quoted-text rule
- For aspect D the quoted text must be taken character-for-character from visible content in the HTML and must appear inside double quotes within the sentence
5 Style constraints
- Do not include any punctuation other than the single period that ends each sentence
- The quoted text may include punctuation exactly as it appears in the HTML but the rest of the sentence must not contain commas semicolons parentheses colons dashes or any other punctuation
- Do not include lists explanations notes or any text outside the list
- Do not output any additional commentary headers or metadata
Now analyze the provided HTML and return the JSON array of 13 sentences.
{html_code}
    \end{lstlisting}
    \end{tcolorbox}
    \caption{Prompt for generating webpage instructions.}
    \label{fig:webpage instruction}
\end{figure*}

\begin{figure*}[t]
    \centering
    \begin{tcolorbox}[
        title=Prompt for background consistency visual content instruction following metric,
        fonttitle=\bfseries,
        colback=blue!5!white,
        colframe=blue!75!black,
        boxrule=1pt
    ]
    \begin{lstlisting}[style=promptstyle]
You are an expert AI image compliance analyst. Your task is to evaluate a set of AI-generated images based on their compliance with a 'Global Instruction' that dictates a specific, consistent background.
You must give your output strictly as a single integer.
**RULES:**
**1. Input:**
A 'Global Instruction' describing a required background and a sequence of images will be provided. The very first image is the original source product photograph. All following images are AI-generated.
**2. Objective:**
Your evaluation must focus on the background across the generated images:
* **Adherence:** How well does each individual background match the description in the 'Global Instruction' (e.g., materials, objects, lighting)?
* **Consistency:** Critically, do the key background elements in the **entire set of generated images** maintain a consistent identity? For example, if a plant is described, it must be the *exact same plant* across all images, not just a similar type. While reasonable variations in the arrangement, position, or camera focus of these elements are acceptable in service of good composition, the core components themselves must be recognizably identical.
**3. Scoring Scale (from 0 to 5):**
Your score must reflect the overall success of the entire generated set in achieving a consistent background.
* **5 (Perfect & Consistent Implementation):** Flawless execution. The background in every generated image perfectly matches the instruction, AND the key background elements are unambiguously identical in identity across the entire set.
* **4 (Excellent Implementation):** The background adheres to the instruction excellently. The key elements maintain a very high degree of identity, with any deviations being trivial and barely noticeable.
* **3 (Partial Implementation):** The general background theme is correct across most images, but the key elements are clearly not identical (e.g., using a similar but different plant in each shot), showing a failure in consistency.
* **2 (Poor Implementation):** A mixed result where only some images adhere to the background instruction, or the implementation is incorrect across the entire set (e.g., a key described element is missing everywhere).
* **1 (Minimal Implementation):** The background is poorly implemented and largely incorrect or inconsistent across most of the set. The core requirements of the instruction are missed.
* **0 (No Implementation):** The background instruction is disregarded in the vast majority of images, showing a near-complete failure.
**4. Output Format:**
Respond with ONLY a single integer (from 0 to 5). No explanation, labels, or punctuation.
**Global Instruction:**
{global_pattern}
**Images to Evaluate:**
[First image is the source product, followed by the generated set]
                \end{lstlisting}
    \end{tcolorbox}
    \caption{Prompt for background consistency visual content instruction following metric.}
    \label{vcif-background}
\end{figure*}

\begin{figure*}[t]
    \centering
    \begin{tcolorbox}[
        title=Prompt for watermark consistency visual content instruction following metric,
        fonttitle=\bfseries,
        colback=blue!5!white,
        colframe=blue!75!black,
        boxrule=1pt
    ]
    \begin{lstlisting}[style=promptstyle]
You are an expert AI image compliance analyst. Your task is to evaluate a set of AI-generated images based on their compliance with a 'Global Instruction' that dictates the addition of a specific, consistent watermark.
You must give your output strictly as a single integer.
**RULES:**
**1. Input:**
A 'Global Instruction' describing a required watermark and a sequence of images will be provided. The very first image is the original source product photograph. All following images are AI-generated.
**2. Objective:**
Your evaluation must focus on the watermark across the generated images:
* **Adherence:** In each image, does the watermark precisely match the instruction's requirements for position, shape, color, transparency, and size?
* **Consistency:** Critically, is the watermark **absolutely identical** across the entire set of generated images? It must be the exact same element in every image, not a newly generated one that just looks similar.
**3. Scoring Scale (from 0 to 5):**
Your score must reflect the overall success of the entire generated set in applying a consistent watermark.
* **5 (Perfect & Consistent Implementation):** Flawless execution. The watermark is perfectly implemented according to the instruction and is absolutely identical and identically placed in every single generated image.
* **4 (Excellent Implementation):** The watermark is implemented excellently, with very high consistency. Any deviations are minor and barely perceptible (e.g., a one-pixel shift in one image, a negligible difference in color shade).
* **3 (Partial Implementation):** The watermark is present in most or all images, but it is inconsistent across the set (e.g., different positions, sizes, or colors).
* **2 (Poor Implementation):** The watermark is correctly applied to only a minority of the images, or it is fundamentally incorrect in its attributes (e.g., wrong shape or color) wherever it appears.
* **1 (Minimal Implementation):** The watermark is poorly implemented and missing from the majority of images.
* **0 (No Implementation):** The watermark instruction is disregarded in the vast majority of images, showing a near-complete failure.
**4. Output Format:**
Respond with ONLY a single integer (from 0 to 5). No explanation, labels, or punctuation.
**Global Instruction:**
{global_pattern}
**Images to Evaluate:**
[First image is the source product, followed by the generated set]
                \end{lstlisting}
    \end{tcolorbox}
    \caption{Prompt for watermark consistency visual content instruction following metric.}
    \label{vcif-watermark}
\end{figure*}

\begin{figure*}[t]
    \centering
    \begin{tcolorbox}[
        title=Prompt for character consistency visual content instruction following metric,
        fonttitle=\bfseries,
        colback=blue!5!white,
        colframe=blue!75!black,
        boxrule=1pt
    ]
    \begin{lstlisting}[style=promptstyle]
You are an expert AI image compliance analyst. Your task is to evaluate a set of AI-generated images based on their compliance with a 'Global Instruction' that dictates the use of the same real-life model.
You must give your output strictly as a single integer.
**RULES:**
**1. Input:**
A 'Global Instruction' specifying the use of a consistent human model and a sequence of images will be provided. The very first image is the original source product photograph. All following images are AI-generated.
**2. Objective:**
Your evaluation must focus on the identity and consistency of the human model:
* **Adherence:** Does each generated image feature a real-life model showcasing the product as instructed?
* **Consistency:** Critically, is the model in the **entire set of generated images** the exact same person? Check for consistency in hair (style, color, texture), skin tone, ethnicity, body type, and other defining characteristics. If the face is visible, pay attention to facial features (eyes, nose, jawline).
**3. Scoring Scale (from 0 to 5):**
Your score must reflect the overall success of the entire generated set in maintaining the identity of a single model.
* **5 (Perfect & Consistent Implementation):** Flawless execution. The model is unambiguously the **exact same person** across all generated images. All relevant key identity features are perfectly consistent.
* **4 (Excellent Implementation):** The model is clearly the same person, but there may be very minor and negligible inconsistencies that can be attributed to lighting or angle. The identity remains strong and convincing.
* **3 (Partial Implementation):** There is a clear attempt at consistency, but the model looks more like a "sibling" across images, similar but with noticeable differences in key features.
* **2 (Poor Implementation):** The model is clearly a different person in about half of the images, showing a significant failure in consistency.
* **1 (Minimal Implementation):** The model is a different person in the vast majority of images. There was little to no attempt at consistency.
* **0 (No Implementation):** The instruction to use a model is ignored, or the images fail completely to show a consistent person across the set.
**4. Output Format:**
Respond with ONLY a single integer (from 0 to 5). No explanation, labels, or punctuation.
**Global Instruction:**
{global_pattern}
**Images to Evaluate:**
[First image is the source product, followed by the generated set]
\end{lstlisting}
    \end{tcolorbox}
    \caption{Prompt for character consistency visual content instruction following metric.}
    \label{vcif-character}
\end{figure*}

\begin{figure*}[t]
    \centering
    \begin{tcolorbox}[
        title=Prompt for perspective coherence visual content instruction following metric,
        fonttitle=\bfseries,
        colback=blue!5!white,
        colframe=blue!75!black,
        boxrule=1pt
    ]
    \begin{lstlisting}[style=promptstyle]
You are an expert AI image compliance analyst. Your task is to evaluate a set of AI-generated images based on their compliance with a 'Global Instruction' that dictates a coherent, sequential change in viewpoint.
You must give your output strictly as a single integer.
**RULES:**
**1. Input:**
A 'Global Instruction' describing a required viewpoint sequence (e.g., rotation, zoom, a left-to-right pan) and a sequence of images will be provided. The very first image is the original source product photograph. All following images are AI-generated.
**2. Objective:**
Your evaluation must focus on the sequential relationship between the images:
* **Adherence:** Does the set of images, when viewed in order, follow the specified viewpoint change?
* **Coherence & Fluidity:** Is the transition of viewpoint between images logical, smooth, and progressive? A sudden, illogical jump in angle breaks the coherence.
**3. Scoring Scale (from 0 to 5):**
Your score must reflect the success of the entire set in creating the specified coherent visual sequence.
* **5 (Perfect & Coherent Implementation):** Flawless execution. The sequence of images perfectly and smoothly demonstrates the instructed viewpoint change. The progression is logical and fluid.
* **4 (Excellent Implementation):** The viewpoint change is correctly implemented and easy to follow, but there may be minor jumps or slight jerkiness between frames. The overall effect is still excellent.
* **3 (Partial Implementation):** The general idea of the viewpoint change is attempted, but the execution is flawed. The sequence might be illogical (e.g., jumping from a front view to a back view then a side view), jerky, or incomplete.
* **2 (Poor Implementation):** The images show different viewpoints, but they lack a clear logical order. The set feels more like a random collection of different angles than a coherent sequence.
* **1 (Minimal Implementation):** Only a couple of images suggest the required sequence, while the rest are random or static. The instruction is largely ignored.
* **0 (No Implementation):** The instruction for a viewpoint change is completely ignored. All images are from the same or random perspectives with no sequential relationship.
**4. Output Format:**
Respond with ONLY a single integer (from 0 to 5). No explanation, labels, or punctuation.
**Global Instruction:**
{global_pattern}
**Images to Evaluate:**
[First image is the source product, followed by the generated set]
                \end{lstlisting}
    \end{tcolorbox}
    \caption{Prompt for perspective coherence visual content instruction following metric.}
    \label{vcif-perspective}
\end{figure*}

\begin{figure*}[t]
    \centering
    \begin{tcolorbox}[
        title=Prompt for webpage instruction following metric,
        fonttitle=\bfseries,
        colback=blue!5!white,
        colframe=blue!75!black,
        boxrule=1pt
    ]
    \begin{lstlisting}[style=promptstyle]
    You will be given:
    1. An instruction that was included in the input when generating HTML code.
    2. The generated HTML code.
    Your task: Determine if the HTML code fully follows the given instruction.
    Rules:
    - If the HTML code clearly and completely follows the instruction, output "yes".
    - If the HTML code fails to follow the instruction, or only partially follows it, output "no".
    - Output exactly one word: "yes" or "no" (lowercase, without punctuation, without extra text).
    Now read the instruction and the HTML, then output only "yes" or "no".
    instruction: {instruction}
    html_code: {html_code}
    \end{lstlisting}
    \end{tcolorbox}
    \caption{Prompt for webpage instruction following metric.}
    \label{wif}
\end{figure*}

\begin{figure*}[t]
    \centering
    \begin{tcolorbox}[
        title=Prompt for webpage design quality metric,
        fonttitle=\bfseries,
        colback=blue!5!white,
        colframe=blue!75!black,
        boxrule=1pt
    ]
    \begin{lstlisting}[style=promptstyle]
You will be shown a single image: a screenshot of a product-display webpage. Evaluate only the static visual information visible in the image (ignore interactivity, performance, accessibility checks that require code, and factual correctness).
Consider the following five criteria in your evaluation:
1) Visual hierarchy & message clarity - Are the main message, headlines, body text, and calls-to-action immediately clear and legible? Is there a distinct visual hierarchy (using size, weight, and style) that guides the user's eye through the content effectively?
2) Layout & spacing - Are content blocks clearly separated? Is there sufficient whitespace for a clean, uncluttered feel? Are column widths and line lengths optimized for comfortable reading?
3) Image sizing & cropping - Are the images sized proportionally within the layout and cropped effectively so that important visual content is fully visible (no awkward edge-cuts or over-cropping), and image scale feels appropriate relative to surrounding elements?
4) Color harmony / palette cohesion - Are the chosen colors harmonious, consistent, and pleasant together?
5) Overall aesthetic appeal - Is the page overall visually attractive, balanced, and polished as a single composition?
Rules:
- Output ONLY a single integer (0-10) and nothing else (no labels, no explanation, no punctuation)
- Use higher scores for visually clear, balanced, and professional-looking designs; use lower scores for cluttered, inconsistent, or visually unpleasant designs.
- Do NOT output 10 for any criterion unless that aspect is near-professional, exemplary quality.
Now evaluate the provided screenshot and respond with just one integer (0-10).
    \end{lstlisting}
    \end{tcolorbox}
    \caption{Prompt for webpage design quality metric.}
    \label{wdq}
\end{figure*}

\begin{figure*}[t]
    \centering
    \begin{tcolorbox}[
        title=Prompt for webpage content appeal metric,
        fonttitle=\bfseries,
        colback=blue!5!white,
        colframe=blue!75!black,
        boxrule=1pt
    ]
    \begin{lstlisting}[style=promptstyle]
You will be shown a single image: a screenshot of a product display webpage. Your task is to evaluate the **page's effectiveness at driving customer interest and purchase intent** - i.e., how likely a customer is to want to buy after viewing this page. Output **one integer from 0 to 10** (inclusive), where 0 = no purchase interest at all and 10 = extremely compelling and likely to convert.
Consider only the persuasive and conversion-related visual and informational cues - ignore factual correctness of product details. Judge based on:
- Clarity of value proposition (is it obvious what the product is and why to buy?)
- Visual emphasis on product and price (prominent images, clear price/discounts)
- Trust & credibility signals (reviews, ratings, guarantees, seller info)
- Call-to-action strength and visibility (CTA label, size, contrast, placement)
- Ease of decision-making (concise benefits, feature clarity, shipping/returns hints)
- Emotional/aspirational appeal and relevance to target audience (imagery, messaging)
- Urgency/ scarcity cues if present (limited-time offers, low-stock indicators)
Scoring rules:
- Output ONLY a single integer (0-10) and nothing else (no labels, no explanation, no punctuation).
- If the page clearly and strongly motivates purchase, use a high score; if it actively discourages purchase, use a low score.
- Do NOT output 10 unless the page is exceptionally persuasive and could be expected to convert at a high rate in real-world conditions.
Now evaluate the provided screenshot and respond with just one integer (0-10).
    \end{lstlisting}
    \end{tcolorbox}
    \caption{Prompt for webpage content appeal metric.}
    \label{wca}
\end{figure*}

\end{document}